\pdfoutput=1
\documentclass[journal]{IEEEtran}
\usepackage{times}
\usepackage{verbatim}
\usepackage{color}
\usepackage{comment}
\usepackage{url}
\usepackage{graphicx}
\usepackage{array}
\usepackage{amsmath,amssymb}
\usepackage{multirow}
\usepackage{multicol}
\usepackage[compatibility=false]{caption}
\usepackage{subcaption}
\usepackage{tabularx}
\usepackage{tabu}
\usepackage{float}
\usepackage{booktabs}
\usepackage{longtable}
\newcolumntype{C}[1]{>{\centering\arraybackslash}p{#1}}
\newcolumntype{M}[1]{>{\centering\arraybackslash}m{#1}}

\usepackage{textcomp}
\usepackage[normalem]{ulem}

\definecolor{green}{rgb}{0, 0.6, 0}

\newcommand{\etal}{et~al.}

\newcommand{\heading}[1]{\noindent\textbf{#1}}

\newcommand{\figref}[1]{Fig.~\ref{fig:#1}}
\newcommand{\tabref}[1]{TABLE~\ref{tab:#1}} 
\newcommand{\eqnref}[1]{Equation~(\ref{eq:#1})}
\newcommand{\secref}[1]{Section~\ref{sec:#1}}

\newcommand{\tb}[1]{\textbf{#1}}

\newcommand{\best}[1]{{\color{red}#1}}
\newcommand{\second}[1]{{\color{green}#1}}

\newcommand{\ignore}[1]{}   

\newcommand{\todo}[1]{ \textcolor{red}{[{\bf TODO}: #1]}}
\newcommand{\torevise}[1]{\textcolor{black}{#1}}
\newcommand{\revision}[1]{\textcolor{black}{#1}}

\begin{document}
\title{Learning Deep Convolutional Networks for Demosaicing}
\author{Nai-Sheng~Syu$^*$, 
		\revision{Yu-Sheng~Chen$^*$},
        Yung-Yu~Chuang
\thanks{$^*$Nai-Sheng~Syu and Yu-Sheng~Chen contributed equally to this work. This work is based on Nai-Sheng~Syu's master thesis \cite{hsu2016thesis}. All authors are with the Department of Computer Science and Information Engineering, National Taiwan University, Taipei, Taiwan, 106. E-mail: {vm3465s939873 $\vert$ \revision{nothinglo} $\vert$ cyy}@cmlab.csie.ntu.edu.tw.}%
}

\markboth{}
{Syu \MakeLowercase{\textit{et al.}}: Learning Deep Convolutional Networks for Demosaicing}

\maketitle

\IEEEpeerreviewmaketitle

\begin{abstract}
This paper presents a comprehensive study of applying the convolutional neural network (CNN) to solving the demosaicing problem. The paper presents two CNN models that learn end-to-end mappings between the mosaic samples and the original image patches with full information. In the case the Bayer color filter array (CFA) is used, an evaluation on popular benchmarks confirms that the data-driven, automatically learned features by the CNN models are very effective and our best proposed CNN model outperforms the current state-of-the-art algorithms. Experiments show that the proposed CNN models can perform equally well in both the sRGB space and the linear space. It is also demonstrated that the CNN model can perform joint denoising and demosaicing. The CNN model is very flexible and can be easily adopted for demosaicing with any CFA design. We train CNN models for demosaicing with three different CFAs and obtain better results than existing methods. With the great flexibility to be coupled with any CFA, we present the first data-driven joint optimization of the CFA design and the demosaicing method using CNN. Experiments show that the combination of the automatically discovered CFA pattern and the automatically devised demosaicing method outperforms other patterns and demosaicing methods. Visual comparisons confirm that the proposed methods reduce more visual artifacts. Finally, we show that the CNN model is also effective for the more general demosaicing problem with spatially varying exposure and color and can be used for taking images of higher dynamic ranges with a single shot. The proposed models and the thorough experiments together demonstrate that CNN is an effective and versatile tool for solving the demosaicing problem.  
\end{abstract}

\section{Introduction}
\label{sec:intro}


\IEEEPARstart{M}{ost} digital cameras contain sensor arrays covered by color filter arrays (CFAs), mosaics of tiny color filters.
Each pixel sensor therefore only records partial spectral information about the corresponding pixel.
Demosaicing, a process of inferring the missing information for each pixel, plays an important role to reconstruct high-quality full-color images\revision{~\cite{gunturk2005demosaicking,li2008systematic,menon2011overview}}.
Since demosaicing involves prediction of missing information, there are inevitably errors, leading to visual artifacts in the reconstructed image. 
Common artifacts include the zipper effects and the false color artifacts.
The former refers to abrupt or unnatural changes of intensities over neighboring pixels while the later is for the spurious colors that are not present in original image.
In principle, the CFA design and the demosaicing method should be devised jointly for reducing visual artifacts as much as possible. However, most researches only focus on one of them.

The Bayer filter is the most popular CFA\revision{~\cite{bayer1976color}} and has been widely used in both academic researches and real camera manufacturing.
It samples the green channel with a quincunx grid while sampling red and blue channels by a rectangular grid. 
The higher sampling rate for the green component is considered consistent with the human visual system.
Most demosaicing algorithms are designed specifically for the Bayer CFA. 
They can be roughly divided into two groups, interpolation-based methods\revision{~\cite{cok1987signal,adams1995interactions,kiku2013residual,
kiku2014minimized,monno2015adaptive,laroche1994apparatus,
adams1997design,kakarala2002adaptive,buades2009self}} and dictionary-based methods\revision{~\cite{mairal2008sparse,mairal2009non}}.
The interpolation-based methods usually adopt observations of local properties  and exploit the correlation among wavelengths.
However, the handcrafted features extracted by observations have limitations and often fail to reconstruct complicated structures.
Although iterative and adaptive schemes could improve demosaicing results, they have limitations and introduce more computational overhead.
Dictionary-based approaches treat demosaicing as a problem of reconstructing patches from a dictionary of learned base patches.
Since the dictionary is learned, it can represent the distribution of local image patches more faithfully and provide better color fidelity of the reconstructed images.
However, the online optimization for reconstruction often takes much longer time, making such methods less practical.



Despite its practical use for decades, researches showed the Bayer CFA has poor properties in the frequency-domain analysis\revision{~\cite{alleysson2005linear}}.
Thus, some efforts have been put in proposing better CFA designs for improving color fidelity of the demosaiced images\revision{~\cite{hirakawa2008spatio, hao2011geometric, bai2016automatic}}. 
Earlier work mainly focused on altering the arrangement of RGB elements to get better demosaicing results in terms of some handcrafted criteria.
Some also explored color filters other than primary colors.
Inspired by the frequency representation of mosaiced images~\cite{alleysson2005linear}, several theoretically grounded CFA designs have been proposed~\cite{hirakawa2008spatio, hao2011geometric}. They however involve considerable human effort.
Recently, automatic methods for generating CFAs have been proposed by exploiting the frequency structure, a matrix recording all the luminance and chrominance components of given mosaiced images~\cite{bai2016automatic}.
However, although theoretically better, most of these CFAs can only reach similar performances as the state-of-the-art demosaicing methods with the Bayer CFA. 
The main reason is that the more complicated CFA designs require effective demosaicing methods to fully release their potential. 
Unfortunately, due to the complex designs, such demosaicing methods are more difficult to devise and, compared with the Bayer CFA, less efforts have been put into developing demosaicing methods for these CFAs. 

We address these issues by exploring the convolutional neural network (CNN). 
Because of breakthroughs in theory and improvements on hardware, recently CNNs have shown promises for solving many problems, such as visual recognition, image enhancement and game playing.
By learning through data, the network automatically learns appropriate features for the target applications.
We first address the demosaicing problem with the popular Bayer CFA (\secref{bayer}).
Inspired by CNN models for super-resolution~\cite{dong2014learning, kim2015accurate}, we present two CNN models, DeMosaicing Convolutional Neural Network (DMCNN, \secref{DMCNN}) and Very Deep DMCNN (DMCNN-VD, \secref{DMCNN-VD}), for demosaicing.
In contrast with handcrafted features/rules by many interpolation-based methods, the CNN models automatically extract useful features and captures high-level relationships among samples.
Experiments show that the CNN-based methods outperforms the state-of-the-art methods in both the sRGB space (\secref{exp_bayer}) and the linear space (\secref{exp_linear}). 
In addition, they could perform denoising and demosaicing simultaneously if providing proper training data.
We next show that the CNN-based methods can be easily adopted for demosaicing with CFA designs other than the Bayer one (\secref{otherCFAs}).
The data-driven optimization approach makes it easy to train the CNN-based demosaicing methods with different CFAs and outperform existing methods (\secref{nonbayer}). 
With its flexibility to be used with any CFA, we present the first data-driven method for joint optimization of the CFA design and the demosaicing method (\secref{CFAdesign}).
Finally, we demonstrate that the CNN-based method can also be applied to solving a more challenging demosaicing problem where the filter array has spatially varying exposure and color (\secref{SVEC}). 
It enables taking images with a higher dynamic range using a shingle shot. 
All together, the paper presents a comprehensive study which thoroughly explores the applications of CNN models to the demosaicing problems. 


\section{Related work}
\label{sec:related}

\heading{Color demosaicing.}
The demosaicing methods can be roughly classified into two categories: {\em interpolation-based}\revision{~\cite{cok1987signal,adams1995interactions,kiku2013residual,
kiku2014minimized,monno2015adaptive,laroche1994apparatus,
adams1997design,kakarala2002adaptive,buades2009self}} and {\em dictionary-based} methods\revision{~\cite{mairal2008sparse,mairal2009non}}. 
\revision{Surveys of early methods can be found in some comprehensive review papers~\cite{gunturk2005demosaicking,li2008systematic,menon2011overview}.}
Recently, Kiku~\etal~\cite{kiku2013residual} proposed a novel way to demosaic images in the residual space, and later extended the method to minimize Laplacian of the residual, instead of the residual itself~\cite{kiku2014minimized}.
The residual space is considered smoother and easier for reconstruction.
Monno~\etal~\cite{monno2015adaptive} proposed an iterative, adaptive version of residual interpolation framework. 

\ignore{
Sparse coding is a representative dictionary-based method. 
It has been shown effective for the related problems of super resolution and denoising.
Mairal~\etal~\cite{mairal2008sparse} showed the first successful attempt for addressing denoising and demosaicing.
A later extension of sparse coding exploited the non-local information within the image~\cite{mairal2009non}.
The CNN-based methods proposed in the paper are highly related to sparse coding methods, but are even more effective and efficient in exploiting complex self-similarity.
}

\heading{CFA design.}
Alleysson~\etal~\cite{alleysson2005linear} analyzed the demosaicing problem in the frequency domain.
In the frequency domain, the CFA pattern is usually decomposed into three components, luminance and two chrominance frequencies.
Hirakawa~\etal~\cite{hirakawa2008spatio} formulated CFA design as a problem to maximize the distances between luminance and chrominance components and obtained the optimal pattern by exhaustive search in the parameter space.
Condat~\cite{condat2011new} followed the same spirit and proposed a CFA design that is more robust to noise, aliasing, and low-light circumstances.
Hao~\etal~\cite{hao2011geometric} and Bai~\etal~\cite{bai2016automatic} each introduced a pattern design algorithm based on the frequency structure proposed by Li~\etal~\cite{li2011frequency}.
Hao~\etal~\cite{hao2011geometric} formulated the CFA design problem as a constrained optimization problem and solved it with a geometric method.
Later, Bai~\etal~\cite{bai2016automatic} introduced an automatic pattern design process by utilizing a multi-objective optimization approach which first proposes frequency structure candidates and then optimizes parameters for each candidate.


\heading{General demosaicing.}
In addition to colors, other properties of light, such as exposures (spatially varying exposure, SVE) and polarization, could also be embedded into the filter array and more general demosaicing algorithms can be used for recovering the missing information. 
Nayar~\etal~\cite{nayar2002assorted} proposed a general demosaicing framework, {\it Assorted Pixel}, by assuming the demosaiced result can be obtained from an $n$-degree polynomial function of neighboring mosaiced pixels.
The whole process can therefore be thought as a regression problem by solving a linear system.
Yasuma~\etal~\cite{yasuma2010generalized} later proposed a more general pattern, {\it Generalized Assorted Pixel}, with the capability to recover monochrome, RGB, high dynamic range (HDR), multi-spectral images while sacrificing spatial resolutions.
We adopt a similar spatially varying exposure and color (SVEC) setting as Nayar~\etal~\cite{nayar2002assorted} to demonstrate the potential of the CNN-based methods for generalized demosaicing.

\heading{Convolution neural networks.}
To date, deep learning based approaches have dominated many high-level and low-level vision problems.
Krizhevsky~\etal~\cite{krizhevsky2012imagenet} showed the deep CNN is very effective for the object classification problem.
In addition to high-level vision problems, CNN is also found effective in many low-level image processing problems,
including deblurring~\cite{SchulerHHS14, xu2014deep}, denoising~\cite{xie2012image, zeng2015dictionary}, super resolution~\cite{dong2014learning, kim2015accurate, kim2015deeply, liao2015video}, colorization~\cite{cheng2015deep}, photo adjustment~\cite{yan2014automatic} and compression artifacts reduction~\cite{dong2015compression}.
Inspired by the successful CNN-based super-resolution methos~\cite{dong2014learning, kim2015accurate}, this paper attempts to address both the demosaicing problem and the CFA design problem using end-to-end CNN models. 

\revision{There were few attemps on applying CNN models to solving the demosaicing problem~\cite{hsu2016thesis, Gharbi2016joint, tan2017deeprl}. 
In SIGGRAPH Asia 2016, Gharbi~\etal~\cite{Gharbi2016joint} proposed a CNN model for joint demosaicing and denoising. It downsamples the mosaic image into a lower-resolution feature map and uses a series of convoultions for computing the residual at the lower resolution. The input mosaic image is then concatenated with the upsampled residual. The final output is constructed by a last group of convolutions at the full resolution and then a linear combination of resultant feature maps. In ICME 2017, Tan~\etal~\cite{tan2017deeprl} proposed a CNN model for Bayer demosaicing. The model first uses bilinear interpolation for generating the initial image and then throws away the input mosaic image. Given the initial image as the input, the model has two stages for demosaicing. The first stage estimates G and R/B channels separately while the second stage estimates three channels jointly. 
Both papers only tackled the Bayer demosaicing problem. On the other hand, this paper addresses a wider set of demosaicing problems, including demosaicing in the linear space, demosaicing with non-Bayer patterns, CNN-based pattern design and demosaicing with a SVEC pattern.}

\section{Demosaicing with the Bayer filter}
\label{sec:bayer}

The Bayer filter is the most popular CFA. In this section, we will focus on demosaicing with the Bayer filter using the convolutional neural network. First, we will introduce two CNN architectures, DMCNN (\secref{DMCNN}) and DMCNN-VD (\secref{DMCNN-VD}), respectively inspired by recent successful CNN models for image super-resolution, SRCNN~\cite{dong2014learning} and VDSR~\cite{kim2015accurate}. 

\subsection{Demosaicing convolutional neural network (DMCNN)}
\label{sec:DMCNN}

\begin{figure}[t]
\centering
\includegraphics[width=\columnwidth]{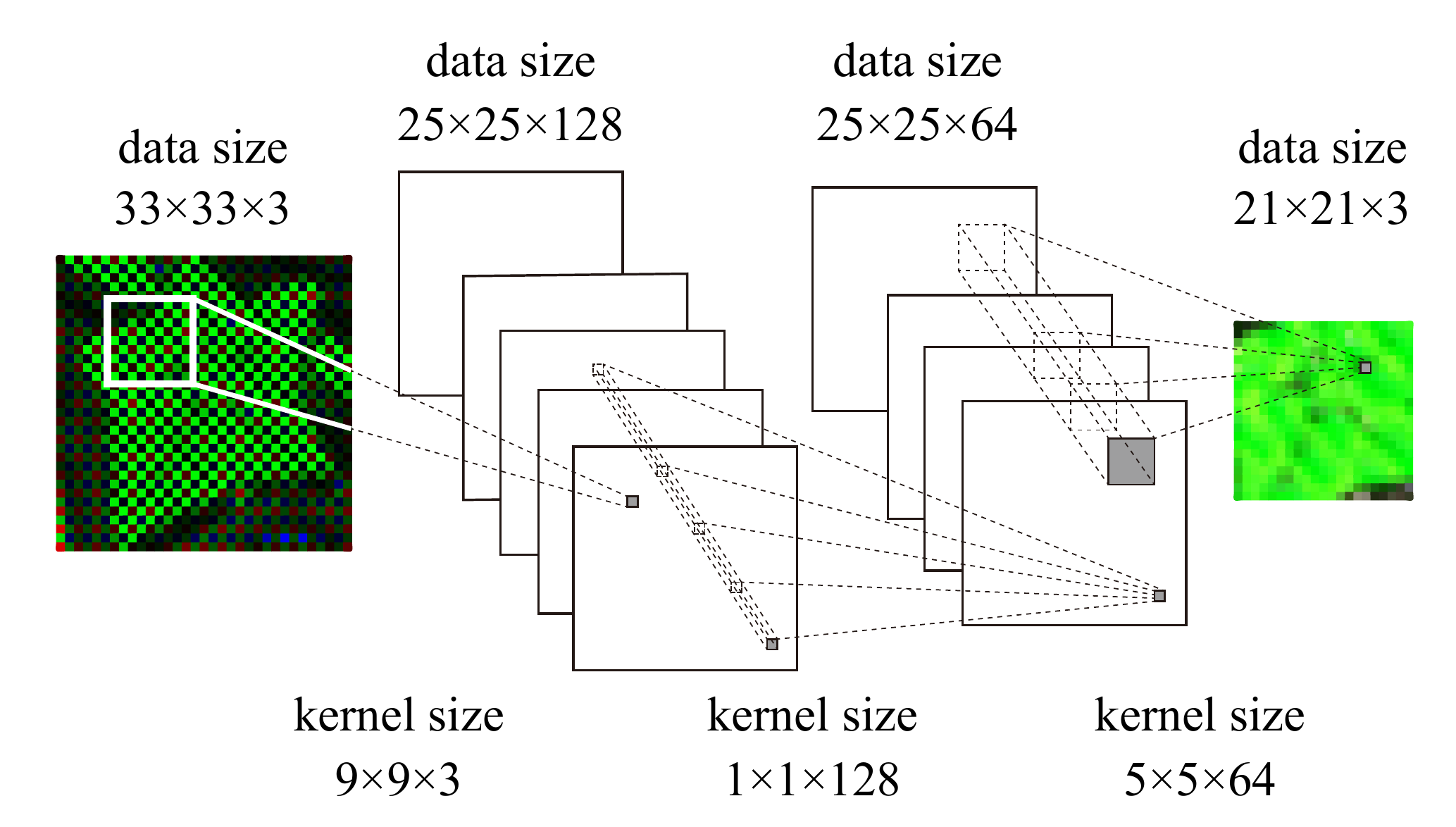}
\caption{
The architecture of DMCNN. The input $33 \times 33$ mosaiced image patch sampled with the Bayer CFA is first extended to $33 \times 33 \times 3$ by adding zeros for missing channels. It then goes through three stages, feature extraction, non-linear mapping and reconstruction. Finally, the network outputs a reconstructed patch with three color channels.}
\label{fig:arch_SRCNN}
\end{figure}

The architecture of the demosaicing convolutional neural network (DMCNN) is inspired by Dong~\etal's SRCNN~\cite{dong2014learning}  for super-resolution. \figref{arch_SRCNN} gives the architecture of DMCNN. 
Since relevant information for demosaicing is often only present locally, patches are densely extracted and presented as the inputs to the network. 
We used $33 \times 33$ patches for DMCNN. Each pixel of the patch consists of three color channels by leaving the two missing channels blank. Thus, the input size is $33 \times 33 \times 3$. Another option would be to have $33 \times 33 \times 1$ patches by using the mosaiced image patch directly. It however could be difficult for the network to figure out which color channel each pixel represents. Four separate networks could be necessary for the four different locations in the Bayer pattern. We found that it is more effective to simply leave missing channel as blank and learn a unified network for different locations of the CFA tile. 
This way, the designed network is also more flexible for different CFA patterns as we will explore in \secref{otherCFAs}.
Similar to SRCNN, DMCNN consists of three layers, each for a specific task:
\begin{enumerate}
\item \textbf{Feature extraction layer.} The first layer is responsible for extracting useful local features. We use 128  $9 \times 9$ filters which are initialized as Gaussian kernels. The output of this layer can be regarded as a low-resolution map of 128-d feature vectors.
\item \textbf{Non-linear mapping layer.} The function of the second layer is to map the extracted high-dimension feature vectors to lower-dimension ones. We use 64 $1 \times 1$ kernels. This way, the non-linear mapping is performed on the pixel itself, without considering the relationships among neighbors.
\item \textbf{Reconstruction layer.} The final layer is designed for  reconstructing a colorful patch from a given set of features. The kernels are $5 \times 5$ and initialized as Gaussian kernels, exploiting the local information to reconstruct the final colors.
\end{enumerate}


ReLU (Rectified Linear Units, $max(0,x)$)~\cite{nair2010rectified} is used as the activation function as it can often avoid the gradient vanishing/exploding problem. Mathematically, the network can be formulated as :
\begin{align}
    F_1(\mathbf{Y}) &= \max(0, W_1 * \mathbf{Y} + B_1 ), \label{eq:DMCNN_F1} \\
    F_2(\mathbf{Y}) &= \max(0, W_2 * F_1(\mathbf{Y}) + B_2), \label{eq:DMCNN_F2} \\
    F(\mathbf{Y}) &= W_3*F_2(\mathbf{Y})+B_3, \label{eq:DMCNN_F3}
\end{align}
where $\mathbf{Y}$ is the input patch; $F_{i}$ is the output feature map of the $i$th layer; $W_{i}$ and $B_{i}$ respectively represent the filters and the bias vector of the $i$th layer; and $*$ is the convolution operator. 
Let $\Theta=\{W_{1},W_{2},W_{3},B_{1},B_{2},B_{3}\}$ denote parameters of the DMCNN network. The $L_2$ norm is used as the loss function.
\begin{align}
    L(\Theta)=\frac{1}{n}\sum_{i=1}^{n}\left\|F(\mathbf{Y}_{i};\Theta)-\mathbf{X}_{i}\right\|^2, \label{eq:DMCNN_loss}
\end{align}
where ${\mathbf{Y}_{i}}$ is the $i$th mosaiced \revision{patch} and ${\mathbf{X}_{i}}$ is its corresponding colorful patch (ground truth); and $n$ is the number of training samples. 
The stochastic gradient descent is used for finding the optimal parameters $\Theta$. 
The learning rate is 1 for the first two layers while a smaller learning rate (0.1) is used for the last layer.

\begin{figure}[t!]
\begin{center}
\begin{tabular}{c}
\includegraphics[width=.99\linewidth]{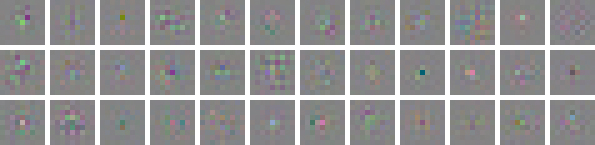} \\
{ (a) learned kernels} \\
\includegraphics[width=.99\linewidth]{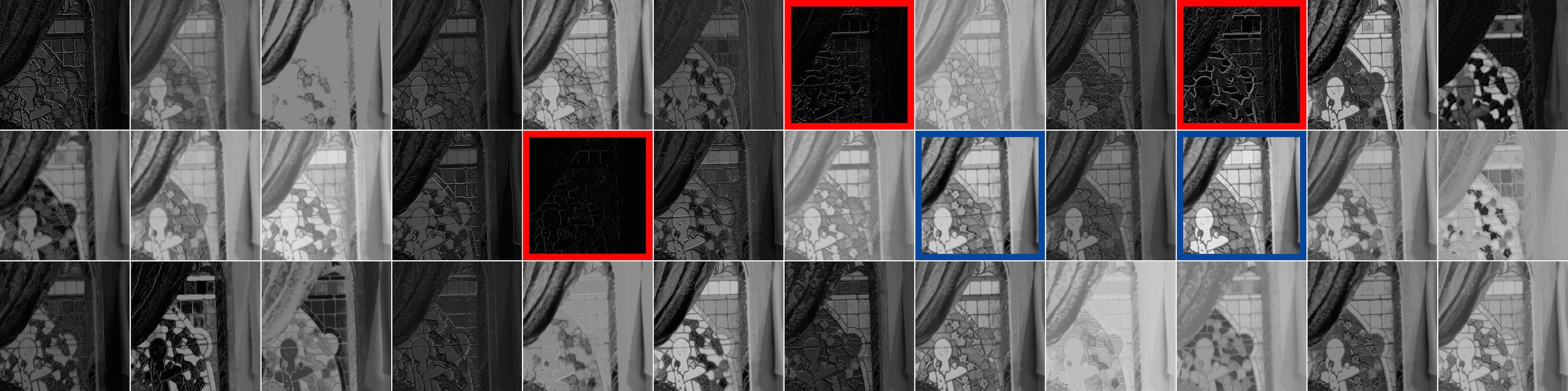} \\
{ (b) feature maps}
\end{tabular}
\end{center}
\caption{Visualization of learned kernels and feature maps for an example. 
(a) 36 of 128 kernels in the first convolution layer.
(b) Corresponding feature maps.}
\label{fig:visualization}  
\end{figure}

The DMCNN network is an end-to-end learning model with two advantages compared to previous demosaicing algorithms. First, the features are explored automatically and optimized in a data-driven manner rather than handcrafted. Second, the reconstruction could exploit more complicated spatial and spectral relationships. 
\figref{visualization} visualizes some learned kernels and corresponding feature maps in the first convolutional layer for an image. 
It can be observed that some automatically learned features explore directional information, which is often considered useful for demosaicing.
For example, the 7th, 10th and 17th features outlined with red in \figref{visualization}(b) are gradient-like features with different orientations.
It can also be found that some features are for chromatic interpolation, such as the 18th and the 20th features outlined with blue in \figref{visualization}(b).
Such features could be difficult to design manually, but can be automatically constructed using CNN.

\subsection{Very Deep DMCNN (DMCNN-VD)}
\label{sec:DMCNN-VD}

Although DMCNN exploits the possibility of learning an end-to-end CNN model for demosaicing, it does not fully explore the potential of CNN models as the model is considerably shallow. It has been shown in many applications that, given the same number of neurons, a deeper neural network is often more powerful than a shallow one. Recently, residual learning has been shown effective on training deep networks with fast convergence~\cite{he2015delving}. 
Residual learning converges faster by learning the residual information and constructing the final solution by adding the learned residual information to the input. 
Kim~\etal~adopted the residual learning approach and proposed a much deeper end-to-end CNN architecture for super-resolution, VDSR~\cite{kim2015accurate}. 
Inspired by their model, we propose a design of a deep CNN model for demosaicing, very deep DMCNN (DMCNN-VD). 

\begin{figure}[t]
\centering
\includegraphics[width=\columnwidth]{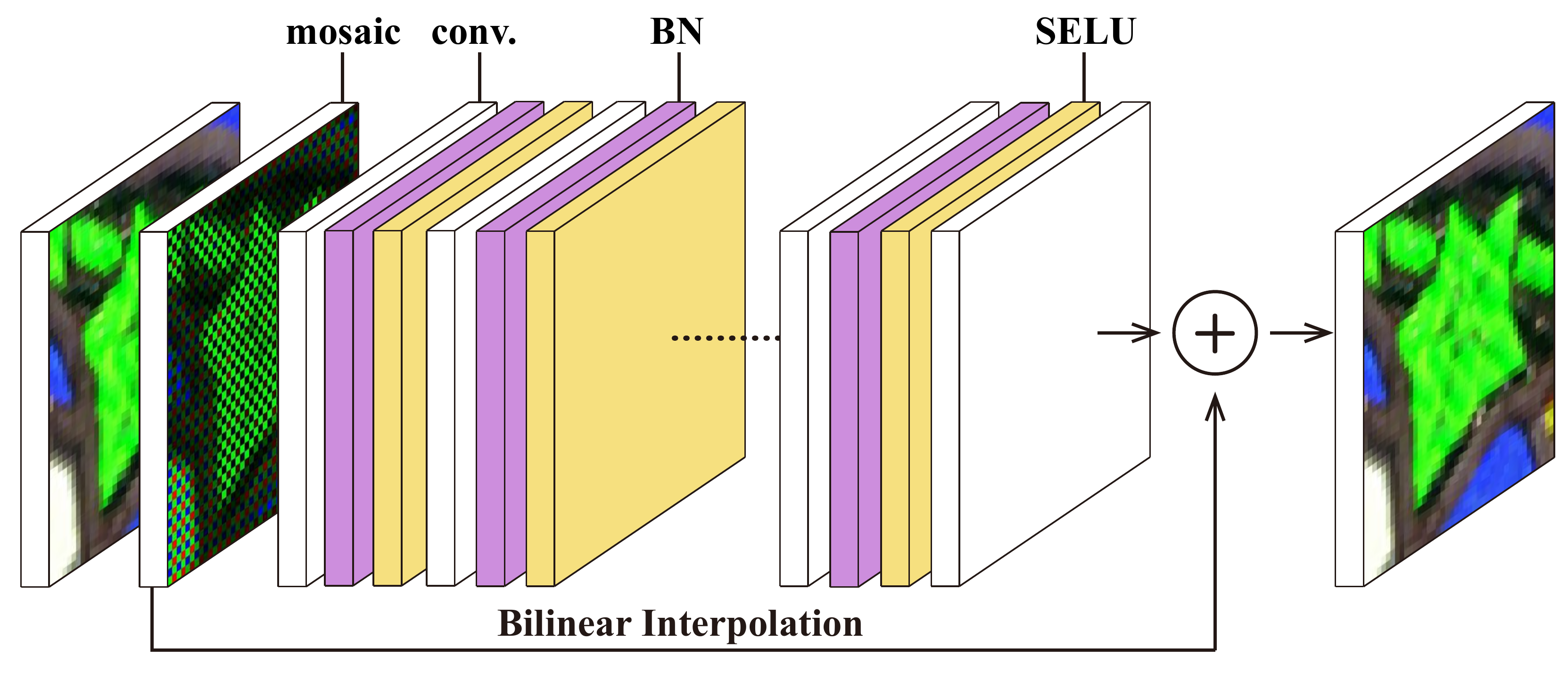}
\caption{\revision{The architecture of DMCNN-VD. It consists of 20 layers with the residual learning strategy. Each layer is composed of a convolution layer, a batch normalization layer and a SELU activation layer.}
}
\label{fig:arch_DMCNN_DR}
\end{figure}

\figref{arch_DMCNN_DR} illustrates the architecture of DMCNN-VD. It consists of $N$ layers ($N\!=\!20$ in our current setting). Each layer is composed of a convolution layer, a batch normalization layer~\cite{ioffe2015bn} and a SELU activation layer~\cite{klambauer2017selu}.
Mathematically, the DMCNN-VD network is formulated as :
\begin{align}
    F_n(\mathbf{Y}) &= selu(W_n * \mathbf{Y} + B_n ), n=1\dots N\!-\!1 
 \label{eq:DMCNN_VD_Fn}\\
    F(\mathbf{Y}) &= W_N*F_{N-1}(\mathbf{Y})+B_N, \label{eq:DMCNN_VD_F}
\end{align}
where $selu(x)=\lambda x$ if $x>0$ and $\lambda \alpha (e^x - 1)$ otherwise. $\lambda$ and $\alpha$ are constants defined in the SELU paper~\cite{klambauer2017selu}. 
\ignore{Note that, different from \eqnref{DMCNN_F2} and \eqnref{DMCNN_F3}, the original input patch is taken as the input to all intermediate layers rather than the output of the previous layer. 
Since the input and output patches are highly correlated, the input patch provides useful information for demosaicing.
If such information is missing, the model has to somehow learn it throughout the network and the information could be difficult to maintain as the architecture goes deeper. By taking the original input to each layer and only learn the residual, the learning process becomes more effective and efficient. }
The loss function is also evaluated using the $L_2$-loss form but with some differences from \eqnref{DMCNN_loss}:
\begin{align}
    L(\Theta)=\frac{1}{n}\sum_{i=1}^{n}\left\|(F(\mathbf{Y}_{i};\Theta)+\tilde{\mathbf{Y}}_{i})-\mathbf{X}_{i}\right\|^2.
\label{eq:residual_loss}
\end{align}
The output of DMCNN-VD, $F(\mathbf{Y}_{i};\Theta)$, refers to the residual between the ground truth patch $\mathbf{X}_{i}$ and \revision{$\tilde{\mathbf{Y}}_{i}$, the bilinear interpolation of the input patch $\mathbf{Y}_{i}$. 
This way, the DMCNN-VD model only focuses on learning the differences between the ground truth and the baseline, often corresponding to the more difficult parts to handle.
Thus, its learning can be more effective and efficient.}
\ignore{Thus, we add $F(\mathbf{Y}_{i};\Theta)$ to the \revision{the bilinear interpolation $\tilde{\mathbf{Y}}_{i}$} and measure its difference from the ground truth. }
\ignore{To have a more correlated input to the ground truth, we use the demosaiced patch by bilinear interpolation as the input.
Thus, $\mathbf{Y}_{i}$ is the bilinear-interpolated demosaiced patch in~\eqnref{residual_loss}.}
In principle, any demosaicing method could be used to generate the input patch.
Although bilinear interpolation could suffer from severe zipper and false color artifacts, it performs as well as the state-of-the-art methods on smooth areas which often represent a large portion of an image.
In addition, as the method is simple, the artifacts tend to be more coherent and the residual information is easier to learn by CNN. Advanced method could produce sophisticated artifacts that are more difficult to learn. 
We found the results of bilinear interpolation are sufficient for residual learning. 
It also has the advantage of being more efficient than other alternatives. 
\ignore{Batch normalization~\cite{ioffe2015bn} is used for each layer and SELU~\cite{klambauer2017selu} is used as the activation function.}

\ignore{
\begin{figure}
\centering
\begin{subfigure}[b]{0.15\linewidth}
\includegraphics[width=\textwidth]{Figures/bilinear_pros/gnd.png}
\caption{}\end{subfigure}
\begin{subfigure}[b]{0.15\linewidth}
\includegraphics[width=\textwidth]{Figures/bilinear_pros/bi.png}
\caption{}\end{subfigure}
\begin{subfigure}[b]{0.15\linewidth}
\includegraphics[width=\textwidth]{Figures/bilinear_pros/cs.png}
\caption{}\end{subfigure}
\begin{subfigure}[b]{0.15\linewidth}
\includegraphics[width=\textwidth]{Figures/bilinear_pros/nls.png}
\caption{}\end{subfigure}
\begin{subfigure}[b]{0.15\linewidth}
\includegraphics[width=\textwidth]{Figures/bilinear_pros/ri.png}
\caption{}\end{subfigure}
\begin{subfigure}[b]{0.15\linewidth}
\includegraphics[width=\textwidth]{Figures/bilinear_pros/mlri.png}
\caption{}\end{subfigure}
\caption{
\torevise{The area that bilinear interpolation can perform well.
(a) ground truth.
(b) bilinear.
(c) CS~\cite{getreuer2012image}.
(d) NLS~\cite{mairal2009non}.
(e) RI~\cite{kiku2013residual}.
(f) MLRI~\cite{kiku2014minimized}.} \todo{could be removed if necessary}
}
\label{fig:bilinear_pros}
\end{figure}
}

\revision{Unless otherwise specified, we used $3 \times 3$ kernels and 1-pixel padding for all the intermediate layers. 
The MSRA initialization policy~\cite{he2015delving} was used for initialization. We used $0.001$ as the factor of standard deviation. Adam~\cite{kingma2014adam} was adopted for gradient updating and we set the learning rate to $1\mathrm{e}{-5}$.}
\tabref{two_arch} gives details for the two proposed demosaicing architectures, DMCNN and DMCNN-VD. 

\begin{table}[]
\centering
\begin{tabular}{|l|c|c|}
\hline
                        & DMCNN    & DMCNN-VD           \\ \hline
Reference architecture  & SRCNN~\cite{dong2014learning} & VDSR~\cite{kim2015accurate} \\ \hline
ConvLayers              & 3        & 20                 \\ \hline
Activation function     & ReLU     & \revision{SELU}               \\ \hline
Kernel size             & 9-1-5    & 3                  \\ \hline
Feature map channel     & 128-64-3 & 64-64-...-3        \\ \hline
Padding zero (pixel)    & 0        & 1                  \\ \hline
Gradient Updating       & Clipping & \revision{Adam}~\cite{kingma2014adam}  \\ \hline
Residual Learning       & No       & Yes                \\ \hline
Initialization          & Gaussian & MSRA~\cite{he2015delving} \\ \hline
\end{tabular}
\caption{Details for the two proposed CNN architectures for demosaicing, DMCNN and DMCNN-VD.}
\label{tab:two_arch}
\end{table}

\begin{table*}[t]
  \centering
  \resizebox{0.8\textwidth}{!}{%
  \begin{tabular}{|c|c|c|c|c|c|c|c|c|c|c|c|c|}
    \hline
    \multirow{3}{*}{} & \multicolumn{4}{c|}{Kodak (12 photos)} & \multicolumn{4}{c|}{McM (18 photos)} & \multicolumn{4}{c|}{Kodak+McM (30 photos)} \\ \cline{2-13}
    Algorithm & \multicolumn{3}{c|}{PSNR} & \multirow{2}{*}{CPSNR} & \multicolumn{3}{c|}{PSNR} & \multirow{2}{*}{CPSNR} & \multicolumn{3}{c|}{PSNR} & \multirow{2}{*}{CPSNR} \\ \cline{2-4} \cline{6-8} \cline{10-12}
    & R & G & B& & R & G & B & & R & G & B & \\ \hline \hline
    SA~\cite{li2005demosaicing} & 39.8 & 43.31 & 39.5 & 40.54 & 32.73 & 34.73 & 32.1 & 32.98 & 35.56 & 38.16 & 35.06 & 36.01 \\ \hline
    SSD~\cite{buades2009self} & 38.83 & 40.51 & 39.08 & 39.4 & 35.02 & 38.27 & 33.8 & 35.23 & 36.54 & 39.16 & 35.91 & 36.9 \\ \hline
    NLS~\cite{mairal2009non} & \second{42.34} & \second{45.68} & \second{41.57} & \second{42.85} & 36.02 & 38.81 & 34.71 & 36.15 & 38.55 & 41.56 & 37.46 & 38.83 \\ \hline
    CS~\cite{getreuer2012image} & 41.01 & 44.17 & 40.12 & 41.43 & 35.56 & 38.84 & 34.58 & 35.92 & 37.74 & 40.97 & 36.8 & 38.12 \\ \hline
    ECC~\cite{jaiswal2014exploitation} & 39.87 & 42.17 & 39 & 40.14 & 36.67 & 39.99 & 35.31 & 36.78 & 37.95 & 40.86 & 36.79 & 38.12 \\ \hline
    RI~\cite{kiku2013residual} & 39.64 & 42.17 & 38.87 & 39.99 & 36.07 & 39.99 & 35.35 & 36.48 & 37.5 & 40.86 & 36.76 & 37.88 \\ \hline
    MLRI~\cite{kiku2014minimized} & 40.53 & 42.91 & 39.82 & 40.88 & 36.32 & 39.87 & 35.35 & 36.60 & 38.00 & 41.08 & 37.13 & 38.32 \\ \hline
    ARI~\cite{monno2015adaptive} & 40.75 & 43.59 & 40.16 & 41.25 & \second{37.39} & \second{40.68} & \second{36.03} & \second{37.49} & \second{38.73} & \second{41.84} & \second{37.68} & \second{39.00} \\ \hline
    PAMD~\cite{zhang2015pamd} & 41.88 & 45.21 & 41.23 & 42.44 & 34.12 & 36.88 & 33.31 & 34.48 & 37.22 & 40.21 & 36.48 & 37.66 \\ \hline
    AICC~\cite{duran2015demosaicking} & 42.04 & 44.51 & 40.57 & 42.07 & 35.66 & 39.21 & 34.34 & 35.86 & 38.21 & 41.33 & 36.83 & 38.34 \\ \hline
    DMCNN & 39.86 & 42.97 & 39.18 & 40.37 & 36.50 & 39.34 & 35.21 & 36.62 & 37.85 & 40.79 & 36.79 & 38.12  \\ \hline
    \revision{DMCNN-VD} & \best{43.28} & \best{46.10} & \best{41.99} & \best{43.45} & \best{39.69} & \best{42.53} & \best{37.76} & \best{39.45} & \best{41.13} & \best{43.96} & \best{39.45} & \best{41.05} \\ \hline
  \end{tabular}
  }
\caption{Quantitative evaluation.
We compared our CNN-based methods (DMCNN and DMCNN-VD) with SA~\cite{li2005demosaicing}, SSD~\cite{buades2009self}, NLS~\cite{mairal2009non}, CS~\cite{getreuer2012image}, ECC~\cite{jaiswal2014exploitation}, RI~\cite{kiku2013residual}, MLRI~\cite{kiku2014minimized}, ARI~\cite{monno2015adaptive}, PAMD~\cite{zhang2015pamd}, AICC~\cite{duran2015demosaicking}. The best method is highlighted in red and the second best is highlighted in \revision{green} in each category (column).}
  \label{tab:psnr_comparison}
\end{table*}

\subsection{Experiments with Bayer demosaicing}
\label{sec:exp_bayer}

\heading{Benchmarks.} The most popular benchmarks for demosaicing are the Kodak dataset and the McMaster dataset. 
All images in the Kodak dataset were captured by film cameras, scanned and then stored digitally.
The dataset contains several challenging cases with high-frequency patterns that are difficult to be recovered from the samples of regular CFA patterns, such as the Bayer pattern.
Zhang~\etal~\cite{zhang2009robust} and Buades~\etal~\cite{buades2009self} noticed that the images in the Kodak dataset tend to have strong spectral correlations, lower saturation, and smaller chromatic gradients than normal natural images.
Thus, Zhang~\etal\ introduced the McMaster benchmark (McM for short) which contains images with statistics closer to natural images~\cite{zhang2011color}.
Since both datasets present their own challenges, demosaicing algorithms are often evaluated on both of them.
We follow the convention used in most of previous studies by using 12 Kodak images and 18 McM images as the evaluation benchmark. 

\heading{Training set.} 
The training data plays an important role in machine learning.
However, we found the training data used in previous demosaicing methods could be problematic. 
For example, the PASCAL VOC'07 dataset was adopted in previous work~\cite{mairal2009non} and it has the following problems:
(1) the images are of low quality which makes some demosaicing artifacts unavoidable, not to mention the compression artifacts in them;
(2) the dataset was collected for object classification and the limited categories of image contents put restrictions, such as the one on the color distribution of images.

For the purpose of training image demosaicing methods, we collected 500 images from Flickr 
with following criteria:
(1) the images are colorful enough to explore the color distributions in real world as much as possible;
(2) there are high-frequency patterns in the images so that CNN learns to extract useful features for challenging cases; and
(3) they are of high quality so that the artifacts due to noise and compression can be avoided as much as possible.
The collected images were resized to roughly 640$\times$480 to have more high-frequency patterns and at the same time, more likely to be mosaic-free.
We call the dataset Flickr500. 
The images were rotated by 90$^{\circ}$, 180$^{\circ}$, 270$^{\circ}$ and flipped in each of directions for data augmentation.
We extracted roughly 3.5 million patches from these images and used them for training the CNN models unless specified otherwise.
The Flickr500 dataset and source codes will be released so that others can reproduce our work\footnote{\url{http://www.cmlab.csie.ntu.edu.tw/project/Deep-Demosaic/}}. 

\heading{Quantitative comparison.}
We quantitatively compare the two proposed CNN models with ten existing algorithms, including SA~\cite{li2005demosaicing}, SSD~\cite{buades2009self}, CS~\cite{getreuer2012image}, ECC~\cite{jaiswal2014exploitation}, AICC~\cite{duran2015demosaicking}, three residual-interpolation-based methods (RI~\cite{kiku2013residual}, MLRI~\cite{kiku2014minimized} and ARI~\cite{monno2015adaptive}) and two sparse-coding-based methods (NLS~\cite{mairal2009non} and PAMD~\cite{zhang2015pamd}).
Following the convention, we use the PSNR (Peak signal-to-noise ratio) value as the metric.
\tabref{psnr_comparison} summarizes the results of the quantitative comparison on the Kodak dataset, the McM dataset and their combination. 
\revision{Note that all numbers in \tabref{psnr_comparison} are directly adopted from previous work~\cite{kiku2013residual, monno2015adaptive} except DMCNN and DMCNN-VD. Thus, we followed the same setting with 12 Kodak images and 18 McM images when obtaining the numbers for DMCNN and DMCNN-VD.}
In each category (a column in the table), the best result is highlighted in red and the second best one in \revision{green}. 
In most cases, we use the CPSNR value on the combined dataset (Kodak+McM) as the final metric. 
The DMCNN model is competitive with the 38.12dB CPSNR value.
However, it is outperformed by the best of ten previous methods, ARI~\cite{monno2015adaptive}, \revision{by} almost 1 dB.
The shallower layers without the residual-learning strategy makes it difficult to recover local details.
On the other hand, with a deeper structure and the residual-learning model, DMCNN-VD reaches \revision{41.05dB} in CPSNR and outperforms all competing algorithms by a  margin, \revision{2.05dB} better than the closest competitor ARI.
\ignore{One thing to note is that, for the Kodak dataset, DMCNN-VD is the best in CPSNR but has a similar performance as NLS.}

\revision{One thing to note is that, both NLS and our methods are learning-based. NLS was trained on the PASCAL VOC 2017 dataset while ours were trained on the Flickr500 dataset. 
To make a fair comparison, we trained DMCNN-VD on the PASCAL VOC 2007 dataset. The CPSNR values are 44.26, 37.35 and 40.11 for Kodak, McM and Kodak+McM respectively while NLS achieves 42.85, 36.15 and 38.83. The DMCNN-VD model still outperforms NLS by a margin.}  
In addition, NLS requires expensive online learning and extra grouping for exploiting sparse coding and self-similarity. Thus, it is less efficient. 
On a PC with an Intel Core i7-4790 CPU and NVIDIA GTX 970 GPU, 
for demosaicing a $500 \times 500$ image, DMCNN took 0.12 second and DMCNN-VD took 0.4 second while NLS took roughly 400 seconds. 
\revision{Note that the CNN models ran with GPUs while NLS only used a CPU. It is not clear how much NLS could be accelerated with parallel computation.}

{\renewcommand{\arraystretch}{2}
\begin{figure*}
  \centering
  \resizebox{2\columnwidth}{!}{%
  \begin{tabular}{C{2.5cm}cccccc}
  \toprule[1pt]
  \multirow{2}{*}{Images} & \multicolumn{1}{C{2.5cm}}{Ground Truth} & \multicolumn{1}{C{2.5cm}}{SSD~\cite{buades2009self}} & \multicolumn{1}{C{2.5cm}}{CS~\cite{getreuer2012image}} & \multicolumn{1}{C{2.5cm}}{NLS~\cite{mairal2009non}} & \multicolumn{1}{C{2.5cm}}{RI~\cite{kiku2013residual}} & \multicolumn{1}{C{2.5cm}}{ARI~\cite{monno2015adaptive}} \\ \cline{2-7}
  & \multicolumn{1}{C{2.5cm}}{DMCNN} & \multicolumn{1}{C{2.5cm}}{DMCNN-VD} & \multicolumn{1}{C{2.5cm}}{DS-VD} & \multicolumn{1}{C{2.5cm}}{CYGM-VD} & \multicolumn{1}{C{2.5cm}}{Hirakawa-VD} & \multicolumn{1}{C{2.5cm}}{DMCNN-VD-Pa} \\ \bottomrule[1pt]
  \end{tabular}
  }
  \renewcommand{\arraystretch}{0.5}
  \begin{tabular}{ccccc}
  &&&&
  \end{tabular}
  \renewcommand{\arraystretch}{3}
  \centering
  \resizebox{2\columnwidth}{!}{%
  \begin{tabular}{ccccccc}
    \includegraphics[width=\linewidth,height=\linewidth,keepaspectratio]{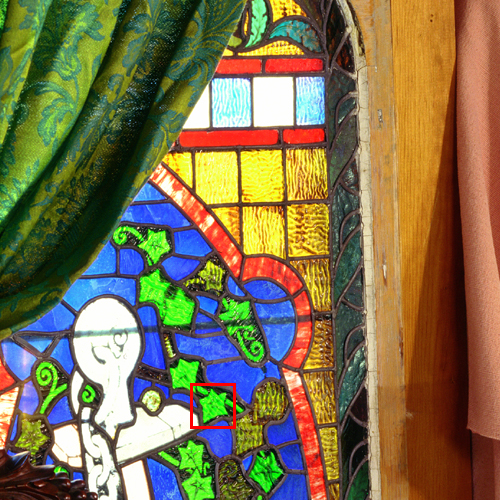} & \includegraphics[width=\linewidth]{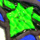} & \includegraphics[width=\linewidth]{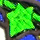} & \includegraphics[width=\linewidth]{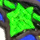} & \includegraphics[width=\linewidth]{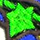} & \includegraphics[width=\linewidth]{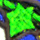} & \includegraphics[width=\linewidth]{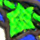} \\
   \includegraphics[width=\linewidth]{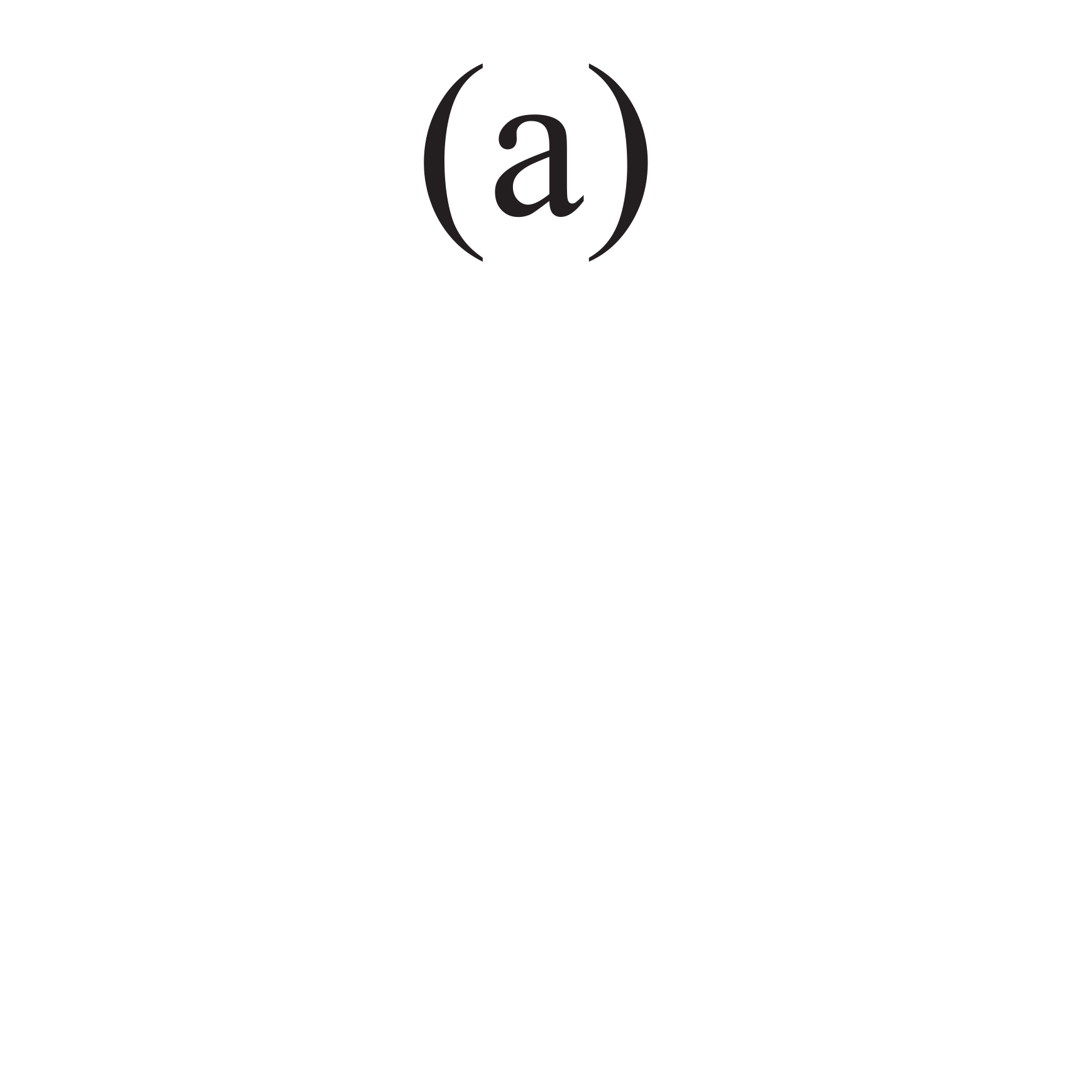}  & \includegraphics[width=\linewidth]{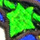} & \includegraphics[width=\linewidth]{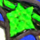} & \includegraphics[width=\linewidth]{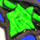} & \includegraphics[width=\linewidth]{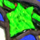} & \includegraphics[width=\linewidth]{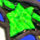} & \includegraphics[width=\linewidth]{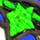} \\
    \includegraphics[width=\linewidth,height=\linewidth,keepaspectratio]{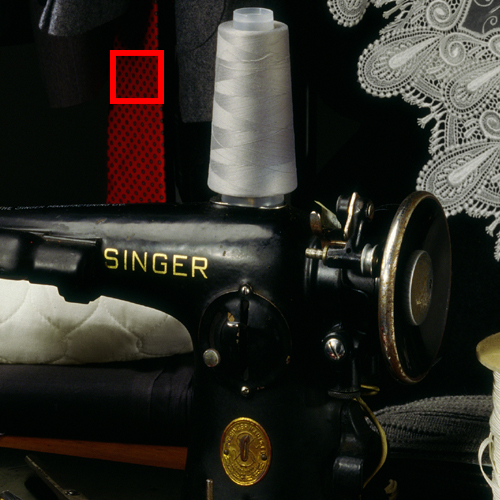} & \includegraphics[width=\linewidth]{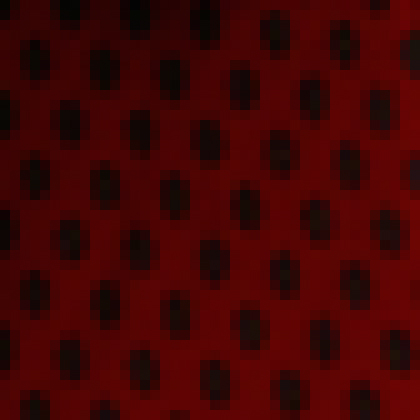} & \includegraphics[width=\linewidth]{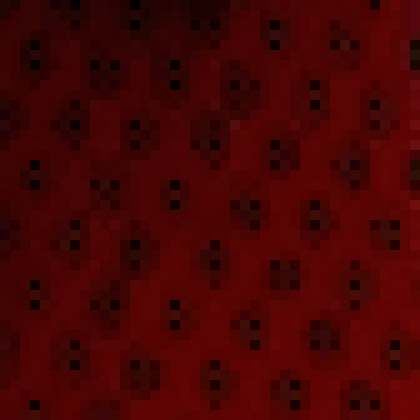} & \includegraphics[width=\linewidth]{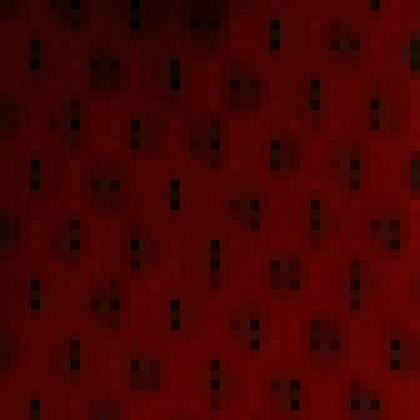} & \includegraphics[width=\linewidth]{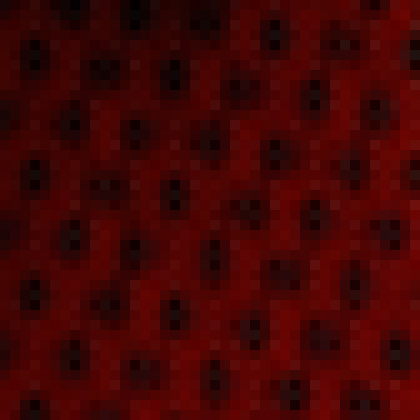} & \includegraphics[width=\linewidth]{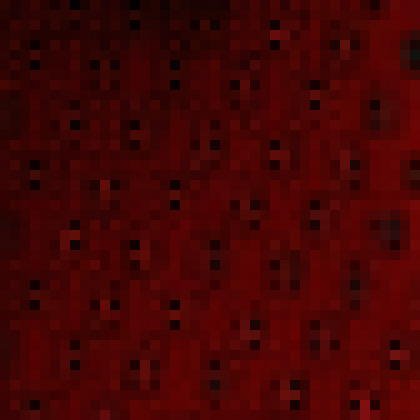} & \includegraphics[width=\linewidth]{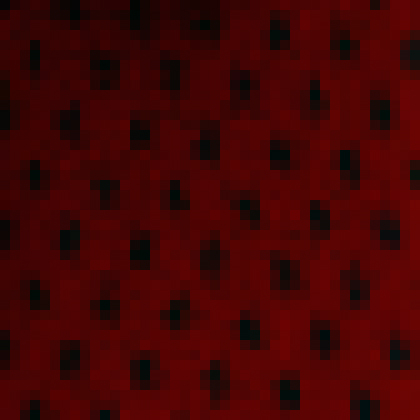} \\
    \includegraphics[width=\linewidth]{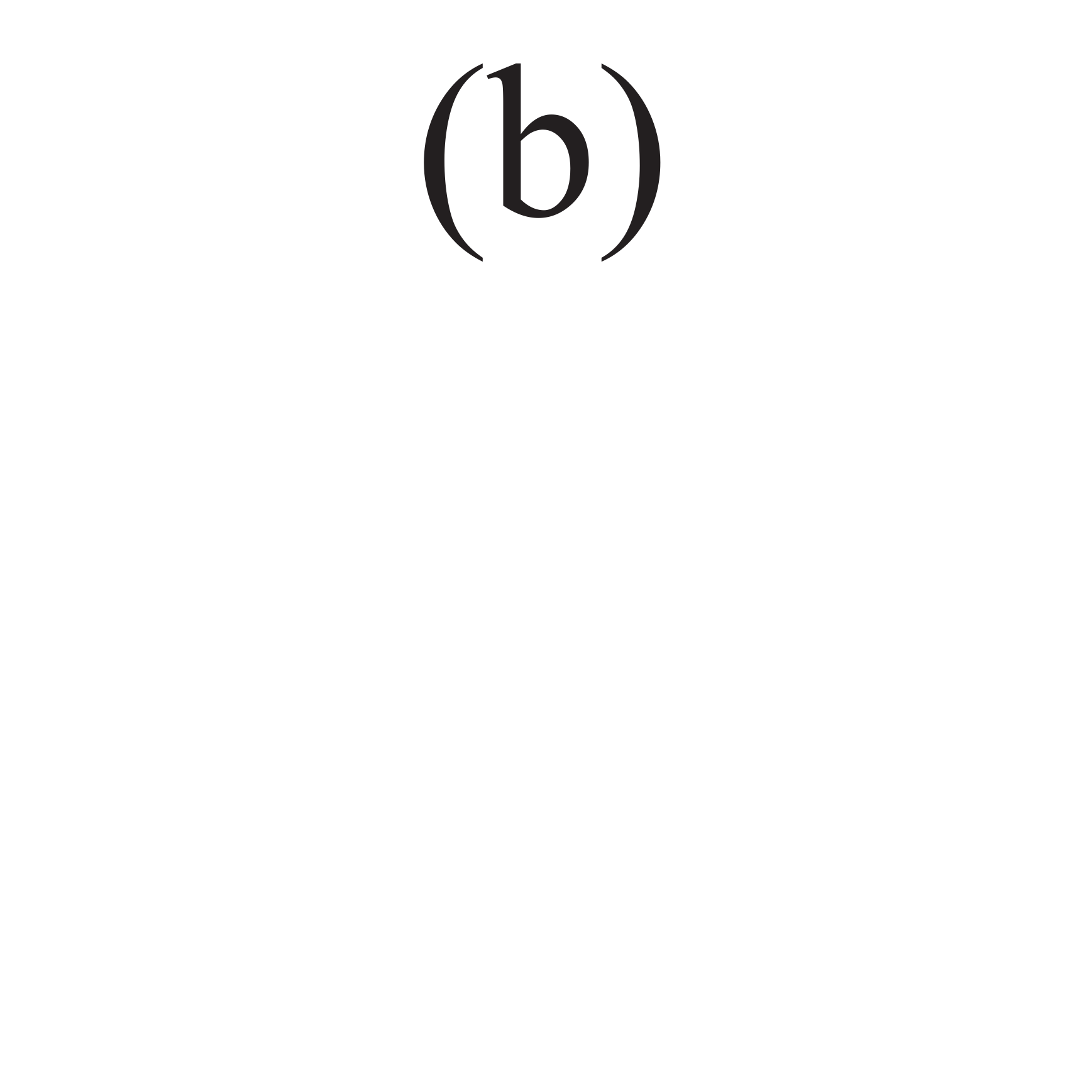} & \includegraphics[width=\linewidth]{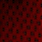} & \includegraphics[width=\linewidth]{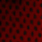} & \includegraphics[width=\linewidth]{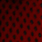} & \includegraphics[width=\linewidth]{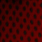} & \includegraphics[width=\linewidth]{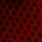} & \includegraphics[width=\linewidth]{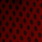} \\
    \includegraphics[width=\linewidth,height=\linewidth,keepaspectratio]{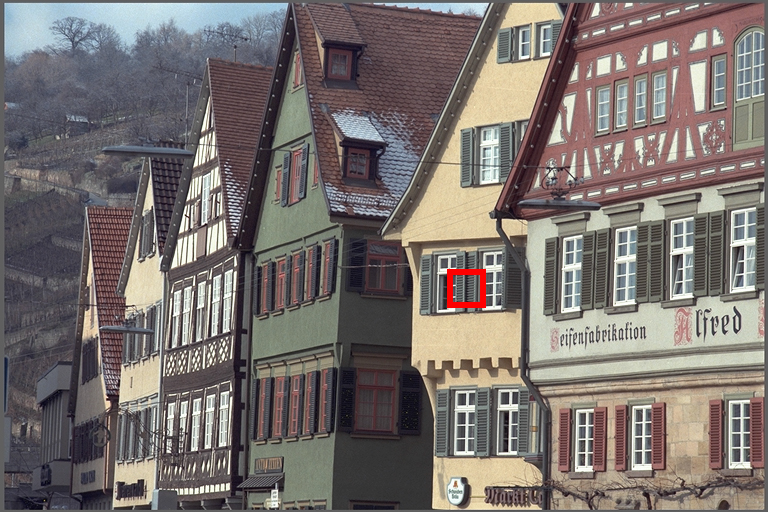} & \includegraphics[width=\linewidth]{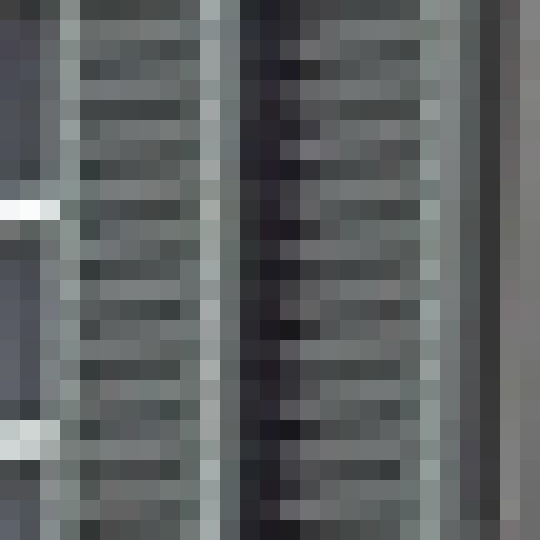} & \includegraphics[width=\linewidth]{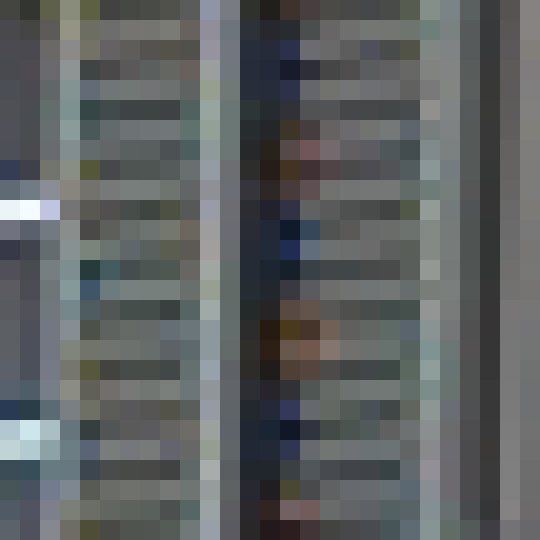} & \includegraphics[width=\linewidth]{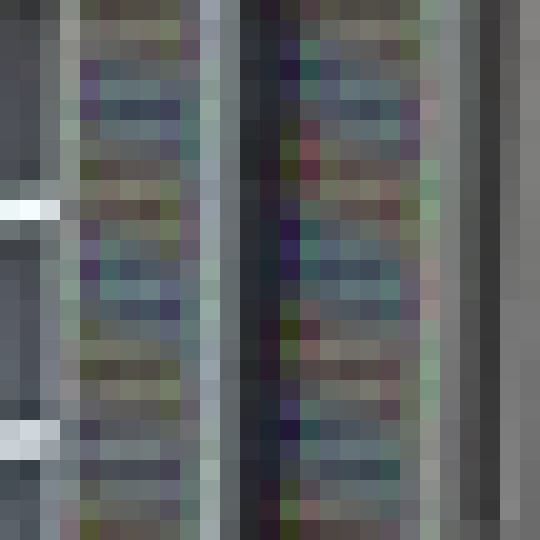} & \includegraphics[width=\linewidth]{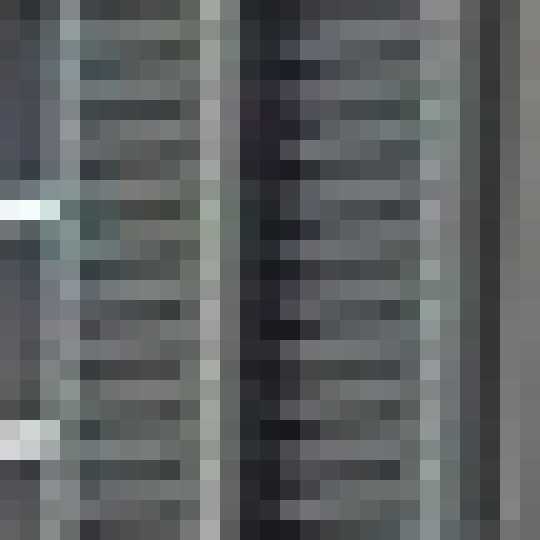} & \includegraphics[width=\linewidth]{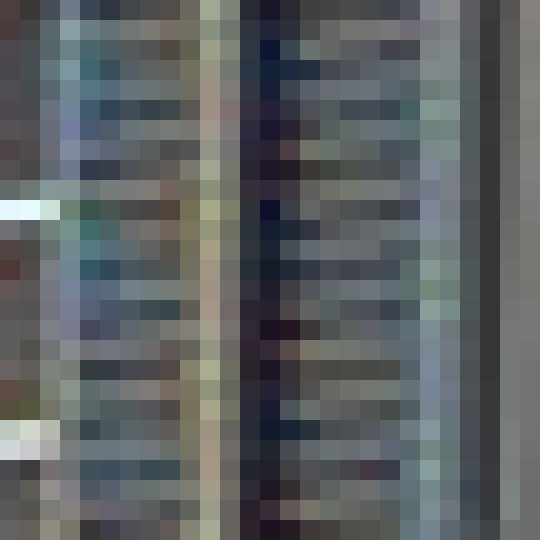} & \includegraphics[width=\linewidth]{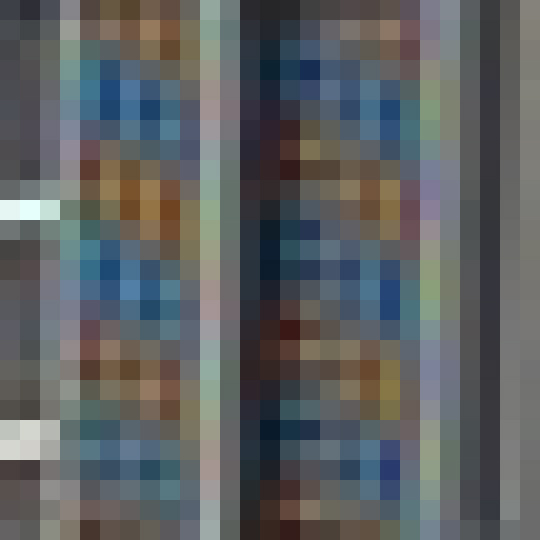} \\
    \includegraphics[width=\linewidth]{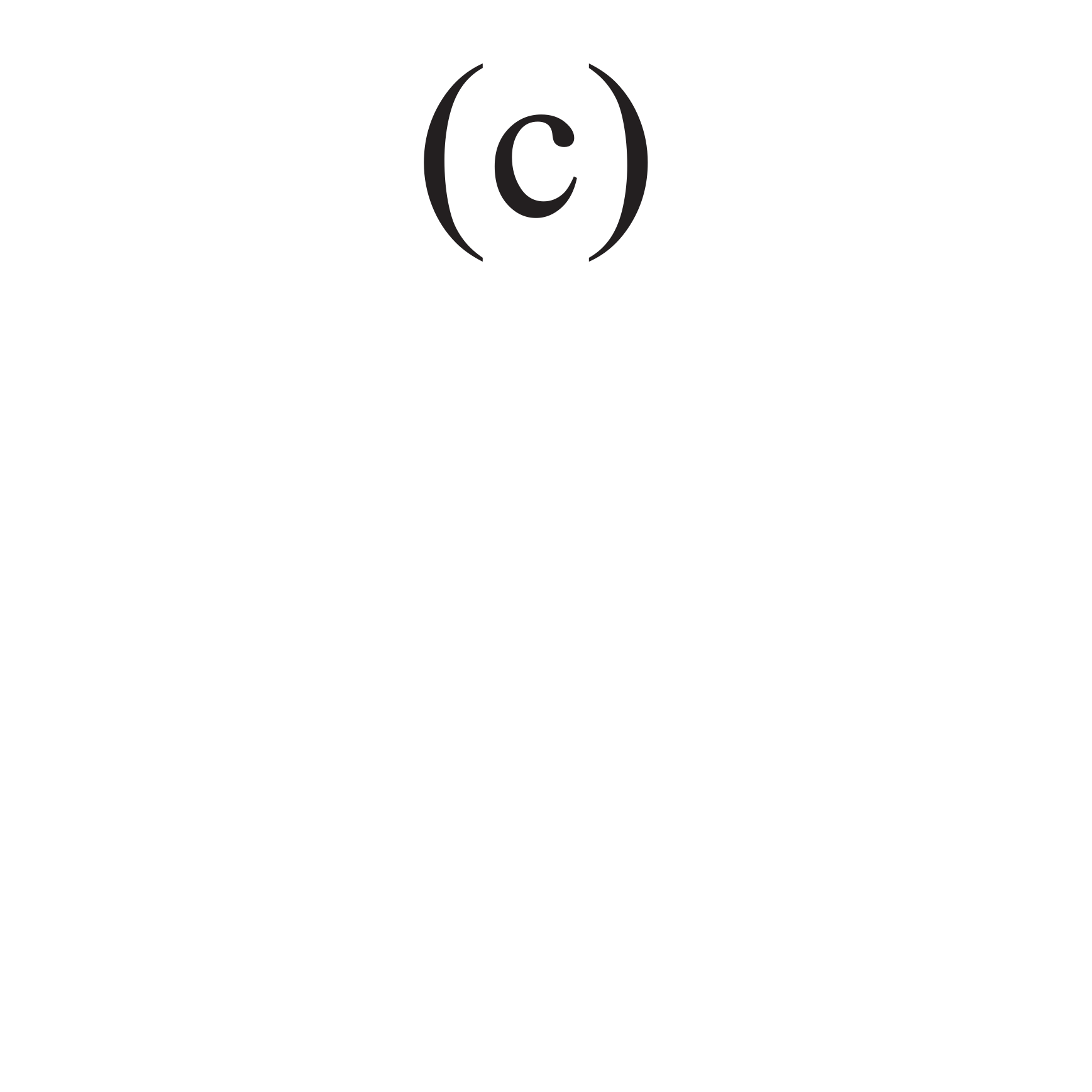} & \includegraphics[width=\linewidth]{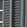} & \includegraphics[width=\linewidth]{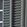} & \includegraphics[width=\linewidth]{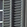} & \includegraphics[width=\linewidth]{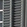} & \includegraphics[width=\linewidth]{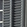} & \includegraphics[width=\linewidth]{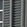} \\
    \includegraphics[width=\linewidth,height=\linewidth,keepaspectratio]{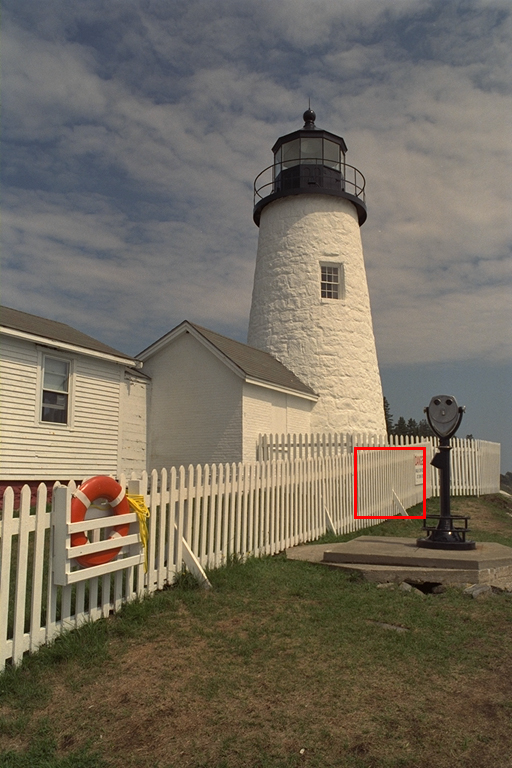} & \includegraphics[width=\linewidth]{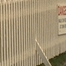} & \includegraphics[width=\linewidth]{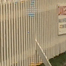} & \includegraphics[width=\linewidth]{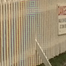} & \includegraphics[width=\linewidth]{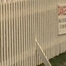} & \includegraphics[width=\linewidth]{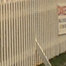} & \includegraphics[width=\linewidth]{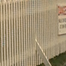} \\
    \includegraphics[width=\linewidth]{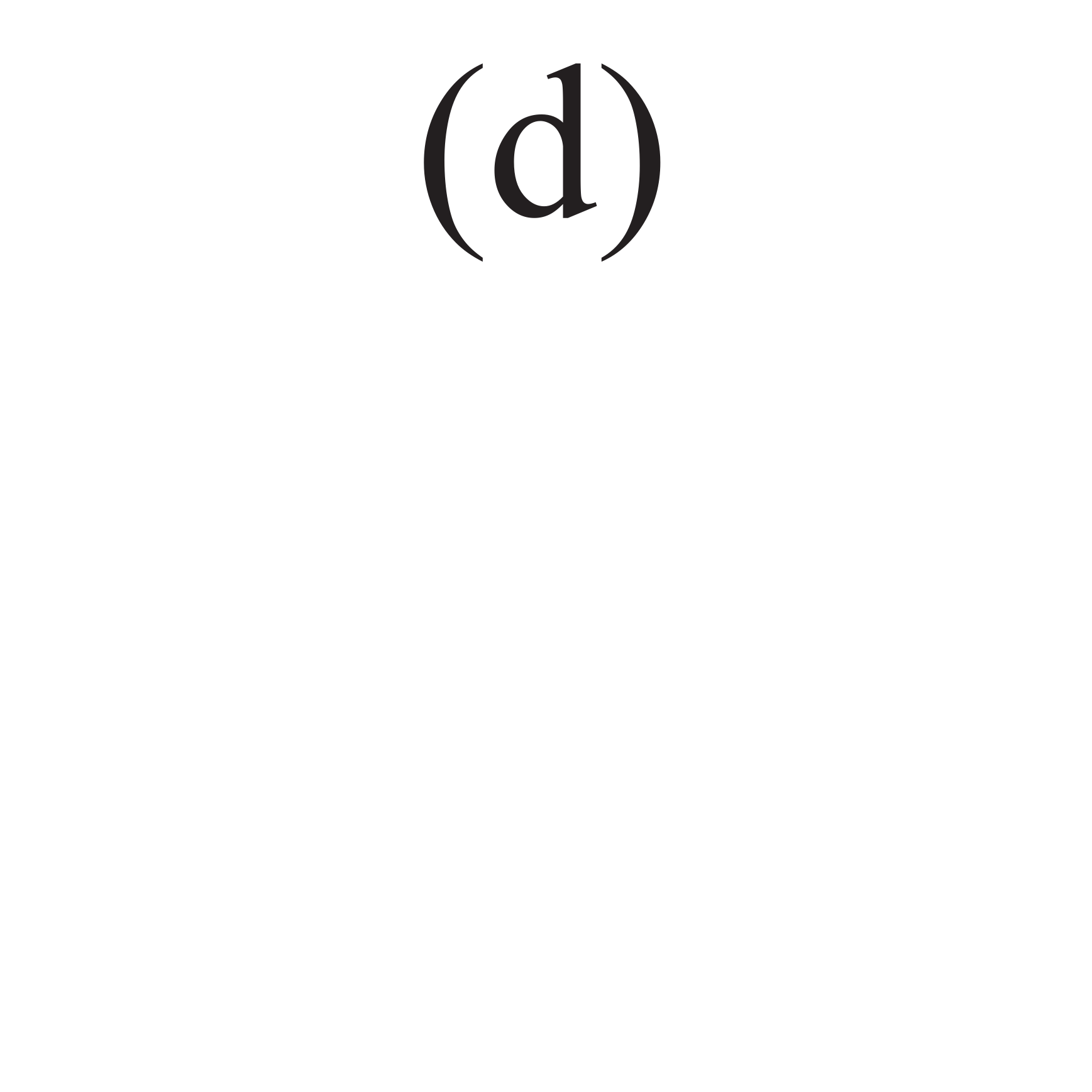} & \includegraphics[width=\linewidth]{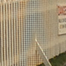} & \includegraphics[width=\linewidth]{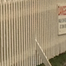} & \includegraphics[width=\linewidth]{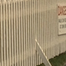} & \includegraphics[width=\linewidth]{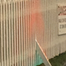} & \includegraphics[width=\linewidth]{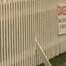} & \includegraphics[width=\linewidth]{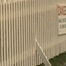}
  \end{tabular}
  }
\caption{Visual comparisons for color demosaicing.  (a)(b) Examples from the McM dataset. (c)(d) Examples form the Kodak dataset. DS-VD, CYGM-VD and Hirakawa-VD are results of DMCNN-VD with diagonal stripe, CYGM and Hirakawa CFAs respectively. DMCNN-VD-Pa represents the DMCNN-VD model that learns CFA design and demosaicing jointly.}
  \label{fig:visual_comp}
\end{figure*}}

\heading{Qualitative comparison.}
\figref{visual_comp} shows visual comparisons on several examples. 
Some models in \figref{visual_comp} will be discussed in \secref{otherCFAs}.
\figref{visual_comp}(a) gives an example from the McM dataset. Most previous methods and the DMCNN model cannot handle such saturated colors and thus produce extra diagonal stripes in the green star.
On the contrary, the DMCNN-VD model performs much better thanks to its deeper architecture through the residual learning scheme.
\figref{visual_comp}(b) shows another example from the McM dataset.
The close-up shows a high-frequency regular pattern, which is difficult to recover for most previous algorithms.
For example, ARI~\cite{monno2015adaptive} generates noisy patterns in this case.
The DMCNN-VD model gives a much better result.
\figref{visual_comp}(c) gives an example from the Kodak dataset.
The close-up shows the blind of the building, containing nearly horizontal stripes.
In this case, residual-interpolated-based methods introduce significant false color artifacts, and so do SSD~\cite{buades2009self} and CS~\cite{getreuer2012image}.
NLS~\cite{mairal2009non} and the DMCNN-VD model have recovered the structure better, showing that such data-driven, automatically learned features can be more effective.
In~\figref{visual_comp}(d), we can observe that the high-frequency structure of the fence is very difficult for all methods to reconstruct perfectly.
Artifacts likes horizontal stripes can be found apparently in the results of interpolation-based methods.
The only successful one is the NLS method~\cite{mairal2009non} which could benefit from its self-similarity strategy.

\begin{table}[t]
  \centering
  \begin{tabular}{|c|c|c|c|}
  \hline
  models                   & Kodak24           & McM & average \\ \hline \hline
  SIGGRAPH Asia~\cite{Gharbi2016joint} 	& 41.20   & 39.50 & 40.47 \\ \hline  
  ICME~\cite{tan2017deeprl} 		   	& 42.04   & 38.98 & 40.73\\ \hline 
  DMCNN-VD  ($3 \times 3$)         	   	& 41.85   & 39.45 & 40.82\\ \hline
  DMCNN-VD ($5 \times 5$)        	   	& 42.19   & 39.56 & 41.06\\ \hline
  DMCNN-VD ($7 \times 7$)        	   	& 42.36   & 39.74 & 41.24\\ \hline    
  DMCNN-VD on WED						&  42.47 & 39.54  & 41.21   \\ \hline
  \end{tabular}
  \caption{\revision{Quantitative comparisons of different deep models on Bayer demosaicing, in terms of the average CPSNR values for the Kodak24 and the McM datasets. Note that, for direct comparison with other CNN models~\cite{Gharbi2016joint, tan2017deeprl}, the setting for the Kodak dataset is different from the one used in \tabref{psnr_comparison}}.}
  \label{tab:comparison_cnn}
\end{table}

\begin{table}[t]
  \centering
  \begin{tabular}{|c|c|c|}
  \hline
  models                   &  100 WED images           &  all WED images \\ \hline \hline
  ICME~\cite{tan2017deeprl} 		   	& 39.67  & - \\ \hline 
  DMCNN-VD on Flickr500        	   	& 40.94  & 40.18\\ \hline
  DMCNN-VD on WED        	   	& 41.55  & -\\ \hline
  \end{tabular}
  \caption{\revision{Quantitative comparisons with the ICME model on the WED dataset. The models were tested on 100 WED images. The DMCNN-VD model trained on the Flickr500 dataset is also tested on all $4,744$ WED images.}}
  \label{tab:comparison_wed}
\end{table}

\heading{Comparisons with other deep demosaicing methods.}
\revision{As mentioned in \secref{related}, there were a couple of prvious papers on deep Bayer demosaicing, one published in SIGGRAPH Asia 2016~\cite{Gharbi2016joint} and the other in ICME 2017~\cite{tan2017deeprl}.
It is difficult to compare with these methods fairly since the training sets are different and the source codes are not always available. 
The SIGGRAPH Asia 2016 model was trained on 2,590,186 $128 \times 128$ difficult patches. 
The ICME 2017 model was trained with 384,000 $50 \times 50$ patches extracted from 4,644 images from the Waterloo Exploration Dataset (WED)~\cite{Ma2017wed}. 
The first two rows of \tabref{comparison_cnn} shows the performance of previous work on the Kodak24 and the McM testing datasets, adopted directly from their papers.
We tested the DMCNN-VD model with the same testing setting as theirs. 
The third row of \tabref{comparison_cnn} reports our results.
With the default kernel size ($3 \times 3$), the DMCNN-VD model has a slight advantage on the average CPSNR value.}  

\revision{The kernel size has impacts on the performance of the model. We experimented with different kernel sizes, $3 \times 3$, $5 \times 5$ and $7 \times 7$, for the DMCNN-VD model. 
\tabref{comparison_cnn} reports the results. It is clear that the performance improves with the kernel size. 
With the $7 \times 7$ kernel, the proposed model achieves the best performance at 42.36dB and 39.74dB for Kodak24 and McM respectively. 
However, a larger kernel size will also incur more computation cost on both training and testing. 
In the paper, without otherwise specified, we report the results with the $3 \times 3$ kernel.} 

\revision{To verify the proposed model with larger datasets, we applied the DMCNN-VD model trained on the Flickr500 dataset to the WED dataset. The WED dataset contains 4,744 images. The DMCNN-VD model achieves 40.18dB in terms of CPSNR. It shows the DMCNN-VD model can generalize very well. In addition, we have trained the DMCNN-VD model using the WED dataset. We used the same setting as the ICME paper in which 4,644 images were used for training and the rest 100 images for testing. The last row of \tabref{comparison_cnn} reports the results. When trained on the same dataset, the DMCNN-VD model achieves 42.27dB and 39.54dB for Kodak24 and McM respectively, outperforming the ICME model\textquotesingle s 42.04dB and 38.98dB. When testing on 100 WED images, the DMCNN-VD (WED) model obtains 41.55dB while the ICME 2017 paper reports 39.67dB.}

\subsection{Experiments with the linear space and noise}
\label{sec:exp_linear}

Like most previous demosaicing papers, the previous section evaluates methods in the sRGB space. 
However, in real camera processing pipeline, the demosaicing process is often performed in the linear space of radiance rather than the sRGB space used in most demosaicing researches.
This issue was recently addressed by Khashabi~\etal~\cite{khashabi2014joint}.
They collected a new dataset called MDD (Microsoft Demosaicing Dataset).
In this dataset, all images were captured by Canon 650D and Panasonic Lumix DMC-LX3.
To simulate mosaic-free images, they proposed a novel down-sampling technique and converted data into the linear space. 
In addition, they also pointed out the input mosaiced images are usually noisy in reality.
As the result, the dataset also provides noisy mosaiced images by adding noise extracted from the original raw images.
In addition, they proposed a method for joint demosaicing and denoising by learning a nonparametric regression tree field (RTF)~\cite{khashabi2014joint}. 
In the following experiments, we will first apply the pre-trained DMCNN-VD model directly to the MDD dataset and then improve its performance by transfer learning.

\heading{Clean data.}
The MDD dataset provides both clean and noisy mosaiced images.
We first experiment with the clean versions for demosaicing.
\tabref{psnr_comparison_mdd_clean} reports the CPSNR values for three methods, ARI~\cite{monno2015adaptive}, RTF~\cite{khashabi2014joint}  and DMCNN-VD, in both the linear space and the sRGB space. 
Since RTF is trained with noisy inputs, it is not surprising that its performance on clean data is not as good as the state-of-the-art algorithm designed for clean inputs, ARI~\cite{monno2015adaptive}.
The proposed DMCNN-VD model performs very well with at least 1dB better than ARI in both spaces.
Note that DMCNN-VD is trained in the sRGB space, but it still manages to perform well in the linear space. 

\begin{table}[t]
  \centering
  \begin{tabular}{|c|c|c|}
  \hline
  Algorithms                   & Linear           & sRGB \\ \hline \hline
  ARI~\cite{monno2015adaptive} & \second{39.94}   & \second{33.26} \\ \hline  
  RTF~\cite{khashabi2014joint} & 39.39            & 32.63 \\ \hline
  \revision{DMCNN-VD}                     & \best{41.35}     & \best{34.78} \\ \hline
  \end{tabular}
  \caption{Quantitative evaluation on the clean data of MDD.
  We compare our DMCNN-VD model to ARI~\cite{monno2015adaptive} and RTF~\cite{khashabi2014joint} by reporting CPSNR values.}
  \label{tab:psnr_comparison_mdd_clean}
\end{table}

\begin{table}[t]
  \centering
  \resizebox{\columnwidth}{!}{%
  \begin{tabular}{|c|c|c|c|c|c|c|c|c|}
  \hline
  \multirow{3}{*}{Algorithms} & \multicolumn{4}{c|}{Panasonic (200)}  & \multicolumn{4}{c|}{Canon (57)} \\ \cline{2-9}
  & \multicolumn{2}{c|}{Linear} & \multicolumn{2}{c|}{sRGB} & \multicolumn{2}{c|}{Linear} & \multicolumn{2}{c|}{sRGB} \\ \cline{2-9}
         & CPSNR & SSIM & CPSNR & SSIM & CPSNR & SSIM & CPSNR & SSIM \\ \hline \hline
  ARI~\cite{monno2015adaptive} & 37.29 & 0.956 & 30.59 & 0.871 & 40.06 & 0.97 & 31.82 & 0.886 \\ \hline
  RTF~\cite{khashabi2014joint} & 37.78 & 0.966 & 31.48 & 0.906 & 40.35 & 0.977 & 32.87 & 0.920 \\ \hline
  \revision{DMCNN-VD}                     & \second{38.33} & \second{0.968} & \second{32.00} &  \second{0.920} & \second{40.99} & \second{0.978} & \second{32.89} & \second{0.927} \\ \hline
  \revision{DMCNN-VD-Tr}                  & \best{40.07} & \best{0.981} & \best{34.08} & \best{0.957} & \best{42.52} & \best{0.987} & \best{35.13} & \best{0.959} \\ \hline
  \end{tabular}
  }
  \caption{Quantitative evaluation on the noisy data of MDD.
  We compare our DMCNN-VD and DMCNN-VD-Tr models to ARI~\cite{monno2015adaptive} and RTF~\cite{khashabi2014joint} by reporting both CPSNR and SSIM.}
  \label{tab:psnr_comparison_mdd_noisy}
\end{table}

\setlength{\tabcolsep}{2pt}
\begin{figure}[t]
\begin{center}
\begin{tabular}{ccccc}
\includegraphics[width=0.18\linewidth]{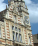} &
\includegraphics[width=0.18\linewidth]{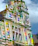} &
\includegraphics[width=0.18\linewidth]{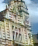} &
\includegraphics[width=0.18\linewidth]{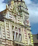} &
\includegraphics[width=0.18\linewidth]{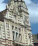} \\
\includegraphics[width=0.18\linewidth]{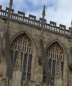} &
\includegraphics[width=0.18\linewidth]{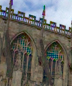} &
\includegraphics[width=0.18\linewidth]{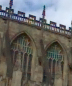} &
\includegraphics[width=0.18\linewidth]{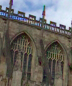} &
\includegraphics[width=0.18\linewidth]{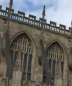} \\
\scriptsize{Ground truth} &
\scriptsize{ARI~\cite{monno2015adaptive}} &
\scriptsize{RTF~\cite{khashabi2014joint}} &
\scriptsize{DMCNN-VD} &
\scriptsize{DMCNN-VD-Tr}
\end{tabular}
\caption{Visual comparisons on MDD examples captured by Panasonic Lumix DMC-LX3 and Canon 650D.}
\label{fig:vis_comp_mdd}
\end{center}
\end{figure}
\setlength{\tabcolsep}{6pt}

\begin{table*}[t]
\centering
\resizebox{0.95\textwidth}{!}{%
\begin{tabular}{|c|c|c|c|c|c|c|c|c|c|c|c|c|c|}
\hline
  \multirow{3}{*}{} & \multirow{3}{*}{} & \multicolumn{4}{c|}{Kodak (12 photos)} & \multicolumn{4}{c|}{McM (18 photos)} & \multicolumn{4}{c|}{Kodak+McM (30 photos)} \\ \cline{3-14}
  Algorithm & Pattern & \multicolumn{3}{c|}{PSNR} & \multirow{2}{*}{CPSNR} & \multicolumn{3}{c|}{PSNR} & \multirow{2}{*}{CPSNR} & \multicolumn{3}{c|}{PSNR} & \multirow{2}{*}{CPSNR} \\ \cline{3-5} \cline{7-9} \cline{11-13}
      &  & R & G & B& & R & G & B & & R & G & B & \\ \hline \hline
NLS~\cite{mairal2009non} & Bayer & 42.34 & 45.68 & 41.57 & 42.85 & 36.02 & 38.81 & 34.71 & 36.15 & 38.55 & 41.56 & 37.46 & 38.83 \\ \hline
ARI~\cite{monno2015adaptive} & Bayer & 40.75 & 43.59 & 40.16 & 41.25 & 37.39 & 40.68 & 36.03 & 37.49 & 38.73 & 41.84 & 37.68 & 39.00 \\ \hline
\revision{DMCNN-VD} & Bayer & \second{43.28} & \best{46.10} & 41.99 & \best{43.45} & 39.69 & \best{42.53} & 37.76 & 39.45 & 41.13 & \best{43.96} & 39.45 & 41.05 \\ \hline
\revision{DMCNN-VD} & Diagonal stripe & 43.16 & 43.69 & \second{42.48} & 43.06 & \second{40.53} & 40.25 & \second{38.70} & 39.61 & \second{41.58} & 41.63 & \second{40.21} & 40.99 \\ \hline
\revision{DMCNN-VD} & CYGM & 40.78 & \second{45.83} & 42.40 & 42.50 & 39.16 & \second{42.33} & \second{38.70} & \second{39.73} & 39.81 & \second{43.73} & 40.18 & 40.84 \\ \hline
\revision{DMCNN-VD} & Hirakawa & 43.00 & 44.64 & 42.43 & \second{43.25} & 40.04 & 40.76 & \second{38.70} & 39.69 & 41.23 & 42.31 & 40.19 & \second{41.12} \\ \hline
Condat~\cite{condat2011new} & Hirakawa & 41.99 & 43.18 & 41.53 & 42.16 & 33.93 & 34.83 & 33.44 & 33.94 & 37.15 & 38.17 & 36.68 & 37.23 \\ \hline
\revision{DMCNN-VD-Pa} & \figref{pre_pattern}(d) & \best{43.42} & 43.80 & \best{42.59} & 43.23 & \best{40.96} & 40.44 & \best{39.02} & \best{39.98} & \best{41.95} & 41.79 & \best{40.45} & \best{41.28} \\ \hline
\end{tabular}
}
\caption{Quantitative comparisons of demosaicing with different CFAs including Bayer CFA, diagonal stripe~\cite{lukac2005color}, CYGM, Hirakawa~\cite{hirakawa2008spatio}, and the proposed data-driven CFA.} 
\label{tab:comp_pattern}
\end{table*}

\heading{Noisy data.}
The noise in the inputs could significantly hurt the performances of demosaicing methods, especially those deriving rules from clean inputs without taking noise into account. 
\tabref{psnr_comparison_mdd_noisy} reports CPSNR and SSIM values in both the linear and sRGB spaces. 
Note that we report the results of Panasonic Lumix DMC-LX3 and Canon 650D separately in \tabref{psnr_comparison_mdd_noisy}  because they have difference noise characteristics. 
It confirms that the algorithm designed for clean data (ARI) could perform less well on noisy inputs.
Although training on clean data, the proposed DMCNN-VD model performs surprisingly as well as the RTF method. 
Since noisy training data are available in MDD, we could leverage them for fine tuning the DMCNN-VD model trained on Flickr500 to improve its performance. 
This can be regarded as a transfer learning strategy~\cite{dong2015compression}.
In our case, the model transfers from the clean sRGB space to the noisy linear space.
We denote the transferred model as DMCNN-VD-Tr.
The CPSNR/SSIM values reported in~\tabref{psnr_comparison_mdd_noisy} show significant improvement by the fine-tuned DMCNN-VD-Tr model.
\figref{vis_comp_mdd} gives a couple of examples for visual comparisons. 
The top row shows an example with the Panasonic camera. 
Due to the noise presented in the input, the results of most algorithms are visually problematic, even the result of RTF shows perceivable color tinting.
Such artifacts are hardly observable in the result of the fine-tuned DMCNN-VD-Tr model.
The bottom row of \figref{vis_comp_mdd} gives an example of the Canon camera. 
Again, the DMCNN-VD-Tr model recovers both color and structure information more faithfully than other methods. 
\figref{linear_visual_comp} shows visual comparisons of more examples on demosaicing in the noisy linear space.  It is clear that the DMCNN-VD-Tr model performs joint demosaicing and denoising well, giving much better results than all other methods.

{
\begin{figure*}
\renewcommand{\arraystretch}{2}
\centering
\resizebox{1.6\columnwidth}{!}{%
  \begin{tabular}{C{3cm}C{3cm}C{3cm}C{3cm}C{3cm}}
  \toprule[1pt]
  Ground truth & ARI~\cite{monno2015adaptive} & RTF~\cite{khashabi2014joint} & DMCNN-VD & DMCNN-VD-Tr \\
  \bottomrule[1pt]
  \end{tabular}
}

\renewcommand{\arraystretch}{0.5}
\begin{tabular}{ccccccccc}
&&&&&&&&
\end{tabular}

\resizebox{1.6\columnwidth}{!}{%
\begin{tabular}{C{3cm}C{3cm}C{3cm}C{3cm}C{3cm}}
\includegraphics[width=\linewidth]{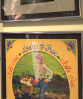}&\includegraphics[width=\linewidth]{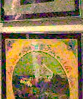}&\includegraphics[width=\linewidth]{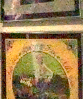}&\includegraphics[width=\linewidth]{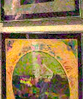}&\includegraphics[width=\linewidth]{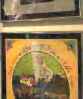}\\
\includegraphics[width=\linewidth]{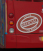}&\includegraphics[width=\linewidth]{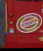}&\includegraphics[width=\linewidth]{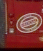}&\includegraphics[width=\linewidth]{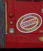}&\includegraphics[width=\linewidth]{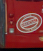}\\
\includegraphics[width=\linewidth]{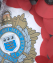}&\includegraphics[width=\linewidth]{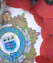}&\includegraphics[width=\linewidth]{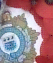}&\includegraphics[width=\linewidth]{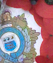}&\includegraphics[width=\linewidth]{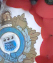}\\
\includegraphics[width=\linewidth]{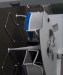}&\includegraphics[width=\linewidth]{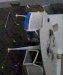}&\includegraphics[width=\linewidth]{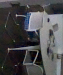}&\includegraphics[width=\linewidth]{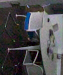}&\includegraphics[width=\linewidth]{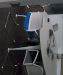}\\
\includegraphics[width=\linewidth]{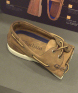}&\includegraphics[width=\linewidth]{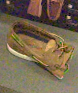}&\includegraphics[width=\linewidth]{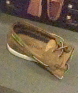}&\includegraphics[width=\linewidth]{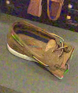}&\includegraphics[width=\linewidth]{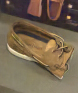}\\
\includegraphics[width=\linewidth]{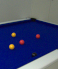}&\includegraphics[width=\linewidth]{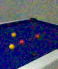}&\includegraphics[width=\linewidth]{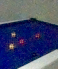}&\includegraphics[width=\linewidth]{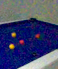}&\includegraphics[width=\linewidth]{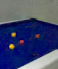}\\
\includegraphics[width=\linewidth]{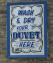}&\includegraphics[width=\linewidth]{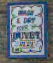}&\includegraphics[width=\linewidth]{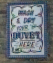}&\includegraphics[width=\linewidth]{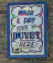}&\includegraphics[width=\linewidth]{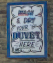}\\
\end{tabular}
}
\caption{Visual comparisons of demosaicing with noisy mosaiced images in the linear space. Both ARI and DMCNN-VD cannot handle noise well since they are trained on clear data. RTF performs better by taking advantages of noisy training data. By transfer learning, the DMCNN-VD-Tr model can perform joint denoising and demosaicing very well. It generates less noisy outputs with much less demosaicing artifacts.}
\label{fig:linear_visual_comp}
\end{figure*}
}

\section{Demosaicing with other CFAs}
\label{sec:otherCFAs}

In this section, we explore CNN models for demosaicing images with CFAs other than the Bayer one. We first apply CNN to demosaicing with three other CFAs (\secref{nonbayer}). Next, we present a data-driven approach for joint optimization of the CFA design and the demosaicing method (\secref{CFAdesign}). Finally, we apply the CNN model to a more challenging demosaicing problem with spatially varying exposure and color (\secref{SVEC}).

\subsection{Demosaicing with non-Bayer CFAs}
\label{sec:nonbayer}

\begin{figure}[t]
\begin{center}
\begin{tabular}{cccc}
\includegraphics[width=0.24\linewidth]{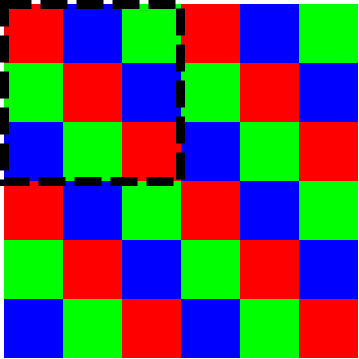} &
\includegraphics[width=0.24\linewidth]{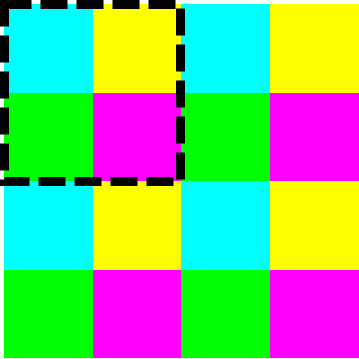} &
\includegraphics[width=0.24\linewidth]{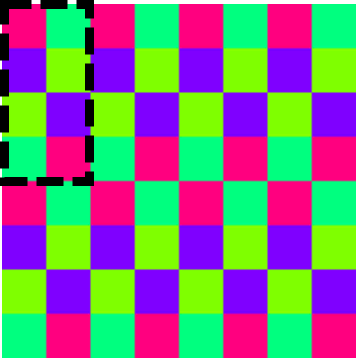} &
\includegraphics[width=0.24\linewidth]{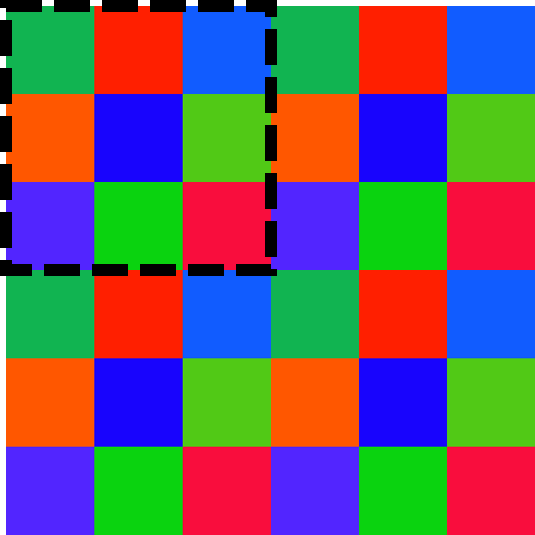} \\
(a) & (b) & (c) & (d)
\end{tabular}
\caption{Examples of different CFA designs: (a) Diagonal stripe~\cite{lukac2005color}, (b) CYGM and (c) Hirakawa~\cite{hirakawa2008spatio} and (d) our CFA design found by the DMCNN-VD-Pa model. 
}
\label{fig:pre_pattern}
\end{center}
\end{figure}

Although the Bayer pattern is the most popular CFA, there are many other CFA designs. 
\figref{pre_pattern} shows three examples, {\em diagonal stripe}~\cite{lukac2005color}, {\em CYGM} and {\em Hirakawa}~\cite{hirakawa2008spatio} CFAs. 
The diagonal stripe CFA (\figref{pre_pattern}(a)) has a $3\times 3$ unit pattern with the three primary colors uniformly distributed.
The CYGM CFA (\figref{pre_pattern}(b)) is proposed as it receives wider range of spectrum than the Bayer pattern.
Its unit pattern is $2\times 2$ with secondary colors and the green color.
Several cameras have been built with this CFA.
Finally, the Hirakawa CFA  (\figref{pre_pattern}(c)) was obtained by optimization through frequency analysis and has a $4\times 2$ unit pattern.

\setlength{\tabcolsep}{2pt}
\setlength{\tabcolsep}{6pt}

Most demosaicing methods are bound up with specific CFAs. They would fail dramatically for other CFAs and often require complete redesigns to work with other CFAs. 
At the same time, most CFAs would require demosaicing methods specifically tailored for them for fully exploring their capability.
One main strength of the demosaicing CNN model is its flexibility. 
The same CNN model can be used for different CFAs as long as it is re-trained with data encoded with the target CFAs.
For a given CFA, the DMCNN-VD model is used while the input layer has to be adjusted with the CFA. As mentioned in \secref{bayer}, the input layer consists of $n$ color planes where $n$ is the number of colors used in the CFA. For the Bayer CFA, three color planes are used because it consists of three primary colors. 
Taking the Hirakawa CFA as an example, its $4 \times 2$ tile consists of four colors, {\em deep pink}, {\em spring green}, {\em slate blue} and {\em chartreuse}.
Thus, four color planes are used. For a pixel sampled with the deep pink channel, the sampled value is filled at the corresponding location of the deep pink color plane while the other three color planes are filled with zeros at the location. 
Three color planes are used for the diagonal stripe CFA and four for CYGM respectively.

\tabref{comp_pattern} reports performances of different combinations of CFAs and demosaicing algorithms. 
The first two rows show the performances of two state-of-the-art methods with the Bayer CFA, NLS and ARI, as the reference.
The next four rows show the performances of the DMCNN-VD model with the Bayer CFA and the three CFAs in \figref{pre_pattern}(a)-(c). 
For each CFA, the DMCNN-VD model is re-trained using the mosaic images sampled with the CFA. 
It is worth noting that, with the learned DMCNN-VD models, the Hirakawa CFA performs the best with \revision{41.12dB}, slightly better than the Bayer pattern with DMCNN-VD. It shows that a better pattern can improve demosaicing performance and the Hirakawa pattern could be the best CFA among the four CFAs experimented.
However, although the Hirakawa pattern seems a better design, it is not easy to release its potential. For example, as shown in the second row from the bottom in \tabref{comp_pattern}, the Hirakawa CFA can only reach a \revision{mediocre} performance at 37.23dB when demosaicing with a previous method, Condat's algorithm~\cite{condat2011new}. 
It reveals that a good CFA design requires a good dedicated demosaicing method to work well. Since fewer methods were devised for CFAs other than the Bayer CFA, their potentials were not fully explored.
The experiment shows how effective and flexible the CNN model is for demosaicing with different CFA designs. 
\figref{visual_comp} shows visual comparisons on several examples. 
DS-VD, CYGM-VD and Hirakawa-VD are results of DMCNN-VD with diagonal stripe, CYGM and Hirakawa CFAs respectively.
\ignore{It is clear that, with a better pattern, the same DMCNN-VD model can perform better. For example, in \figref{visual_comp}(b), the structure is better recovered when using DMCNN-VD with the Hirakawa CFA (Hirakawa-VD) than the Bayer CFA (DMCNN-VD).
Similarly, in \figref{visual_comp}(d), the artifacts of DMCNN-VD on the fence can be removed when the Hirakawa CFA is used. }

\ignore{
Images of this experiment are shown in the bottom row for each example.
The aforementioned difficult case (\figref{visual_comp}d) is now visually pleasing and the PSNRs are still competitive.
We can conclude that the well-designed-CFA can indeed helps us to improve demosaicing results.
As the result, the pattern layer we discussed in~\secref{pattern_overview} might be helpful once it is embedded into CNN so that demosaicing and pattern-designing can be optimized jointly.
}

\subsection{Data-driven CFA design}
\label{sec:CFAdesign}

From the previous section, we learn that the CFA design and the demosaicing algorithm have strong relationship and influence with each other. 
However, to the best of our knowledge, most demosaicing researches focus on either designing mosaic CFAs or devising demosaicing methods and there is no previous work that optimizes both jointly and simultaneously.
Since the CNN model is effective and flexible on learning demosaicing algorithms for various CFAs, it is possible to embed the pattern design into the CNN model to simultaneously learn the CFA design and its demosaicing algorithm by joint optimization. 
The architecture is similar to an autoencoder which finds an effective and compact representation (encoding) for reconstructing the original image faithfully. 
In our case, the representation is formed by spatial color sampling/blending.

\begin{figure}[t]
\centering
\includegraphics[width=0.90\linewidth]{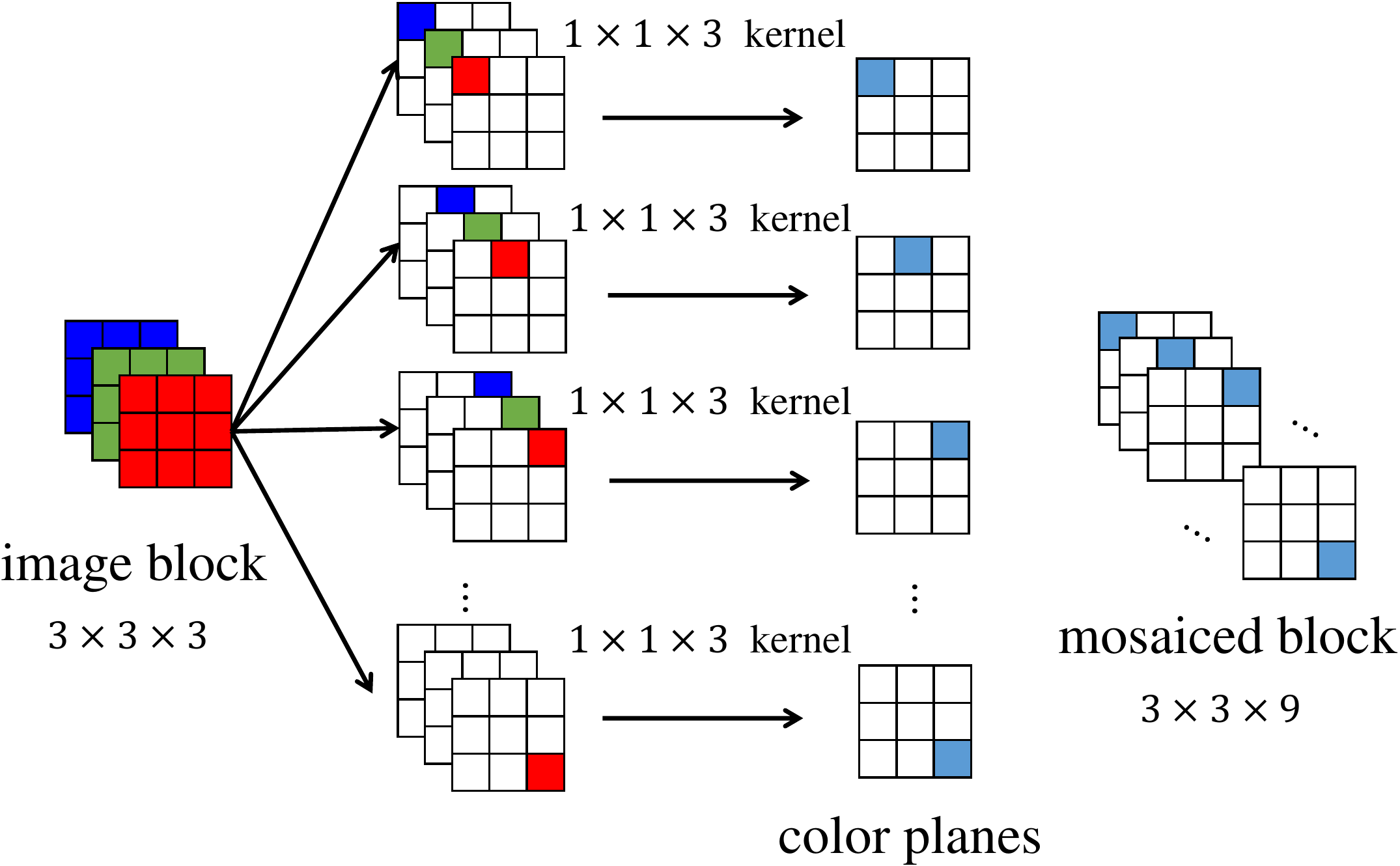}
\caption{The pattern layer. Assuming a $3 \times 3$ unit tile, we need nine color planes, each for a specific location. The layer converts an input $3 \times 3 \times 3$ patch into a $3 \times 3 \times 9$ patch which will be used as the input to the following DMCNN-VD model.}
\label{fig:patternImpl}
\end{figure}

\heading{The pattern layer.}
We first introduce the pattern layer for forming a mosaic pattern. It is different from the popular CNN layers, such as convolution and pooling,  available in deep learning frameworks. 
It cannot be composed using existing layers either. Thus, it has to be implemented as a new layer.  
Assume that the unit pattern is $m \times n$. That is, the unit pattern has $m \times n$ cells and each cell contains a color filter to convert an RGB color into a value of some color channel. We can take the color filter as a $1 \times 1 \times 3$ filter kernel in the CNN model. Thus, we have to learn $mn$ $1 \times 1 \times 3$ kernels for a CFA design. Taking the Bayer CFA as an example, its unit pattern is $2 \times 2$ with four kernels (1, 0, 0), (0, 1, 0), (0, 0, 1), (0, 1, 0) for $\mathbf{R}$, $\mathbf{G}_1$, $\mathbf{B}$, $\mathbf{G}_2$.

\figref{patternImpl} shows an example of the pattern layer with a $3 \times 3$ unit pattern in the forward propagation pass. 
The input is a $3 \times 3 \times 3$ patch. 
For each cell, we have have $1 \time 1 \times 3$ filter to convert its RGB color into a value and put it on the corresponding cell. 
Since we have nine filters, the output consists nine color planes, each corresponding \revision{to} a specific filter. 
Thus, the output of the pattern layer is $3 \times 3 \times 9$. 
Similar to the input structure used in DMCNN and DMCNN-VD models, each output pixel has nine channels. Among them, one is sampled and the other eight are left blank.

\begin{figure}[t]
\centering
\includegraphics[width=\linewidth]{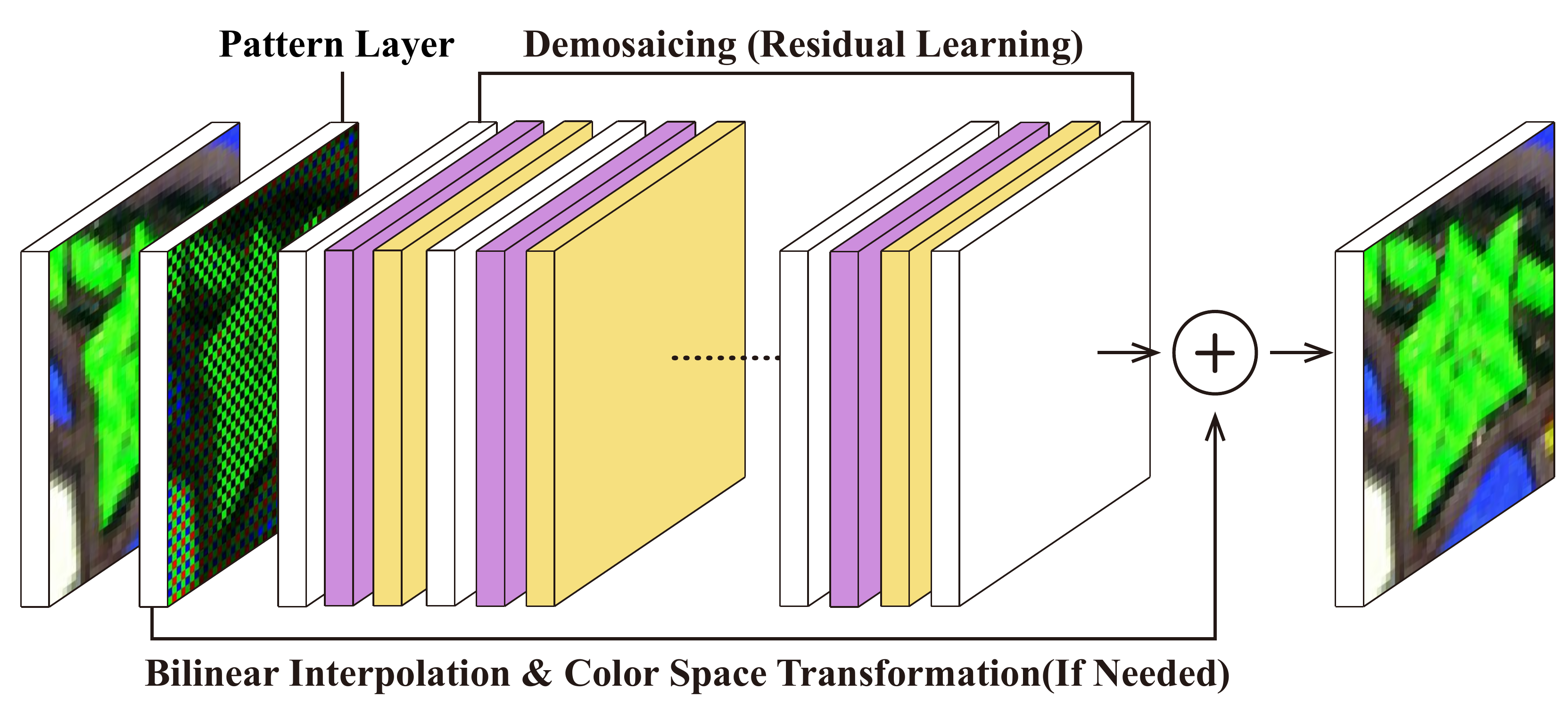}
\caption{\revision{The architecture of the DMCNN-VD-Pa model consisting of two stages: the pattern layer and the DMCNN-VD model. The baseline for residual learning is formed by bilinear interpolation and color transformation.}}
\label{fig:arch_DMCNN_DR_Pa}
\end{figure}

\heading{The DMCNN-VD-Pa model.}
We denote the CNN model with joint optimization of the mosaic pattern and the demosaicing algorithm by DMCNN-VD-Pa.
\figref{arch_DMCNN_DR_Pa} illustrates the DMCNN-VD-Pa model consisting of \revision{two} major components. 
\begin{enumerate}
\item {\em Pattern learning.}
A pattern layer is responsible for learning color filters in CFA.
In the forward pass, the pattern layer sub-sampled the full-color patch (ground truth) using its current filter kernels and outputs a multi-channel mosaiced patch.
In the backward pass, gradients of the kernels are computed as normal convolution layers.
\item {\em Demosaicing.}
The output of the pattern layer, nine color planes for a image patch with 8/9 of information missing, is used as the input to the demosaicing network.
The DMCNN-VD model is used here for demosaicing. 
Note that, since the demosaicing network predicts missing information in all color planes, the output consists nine color planes with full information. 
\end{enumerate}
\revision{Assume that the nine kernels of the $3 \times 3$ unit tile are $\mathbf{C}_1, \mathbf{C}_2, \cdots \mathbf{C}_9$, each representing a RGB color. For residual learning, we first use bilinear interpolation to fill up each color plane. Thus, each pixel now has nine coefficients $\alpha_1, \cdots, \alpha_9$, each for a color plane. 
We then transform the nine coefficients to a RGB color $\mathbf{c}$ by solving a linear system $\mathbf{A} \mathbf{c}= \mathbf{b}$ where $\mathbf{A}$ is a $9 \times 3$ matrix formed by stacking $\mathbf{C}_i$ row by row and $\mathbf{b}$ is the column vector $(\alpha_1, \cdots, \alpha_9)^T$. 
The resultant RGB image is then used as the baseline for residual learning.}

For training the above autoencoder-like CNN model, a set of images are taken as both inputs and labels.
One thing to note is that the optimized pattern could include negative weights in the convolution kernels. Although optimal mathematically, kernels with negative weights are less practical for manufacturing. 
In addition, we would also like to limit the weights so that they are less than 1. 
Unfortunately, constrained optimization for CNN models is difficult. 
Similar to Chorowski and Zurada~\cite{chorowski2015learning}, we adopt the projected gradient descent algorithm~\cite{calamai1987projected} which projects gradients onto the feasible space for each update. 
In our case, for an optimal weight $\tilde{w}_i$ found by regular gradient decent, we update the weights $w_{i}$ of the CNN model as
\begin{align}
   w_{i} = 
    \left\{\begin{matrix}
    0 & \text{if } \tilde{w}_{i}<0\\
    1 & \text{if } \tilde{w}_{i}>1\\
    \tilde{w}_{i} & \text{otherwise.}
    \end{matrix}\right.
    \label{eq:proj_grad}
\end{align}
The weights are initialized with random numbers within $[0,1]$ so that they start with a feasible solution.
\figref{pre_pattern}(d) shows the learned $3 \times 3$ pattern with a couple of interesting properties: (1) the pattern contains primary-color-like lights and (2) the arrangement of cells is regular and similar to the diagonal stripe pattern. \revision{It is worth noting that these properties are related to the chosen size of the unit pattern, $3 \times 3$ . For different sizes of unit patterns, the best pattern could have different characteristics. Exploration with different pattern sizes is left as the future work.}

\heading{Quantitative comparison.}
The last row of \tabref{comp_pattern} shows the performance of DMCNN-VD-Pa on the demosaicing benchmark. 
Its CPSNR value is \revision{41.28dB} on the combined dataset, more than 2.0dB better than ARI~\cite{monno2015adaptive} and \revision{0.23dB} higher than the DMCNN-VD model with the Bayer CFA. 
\revision{The DMCNN-VD model with the Hirakawa pattern is the runner-up with 41.12dB.} 
\ignore{DMCNN-VD-Pa performs better on the newer McM dataset while DMCNN-VD with the Hirakawa pattern does better on the older Kodak dataset.} 
Note that the unit pattern of the Hirakawa pattern is $4 \times 2$ while the DMCNN-VD-Pa's is of $3 \times 3$. 
It is also possible to use the proposed method for finding a good pattern with different tile sizes.
Another interesting thing to note is that DMCNN-VD-Pa performs worse than DMCNN-VD on the green channel. 
It is reasonable since the Bayer CFA has doubled the samples in the green channel.
By contrast, DMCNN-VD-Pa tends \revision{to} sample three channels equally since the $L_2$ loss function simply averages over color channels. 
Since human is more sensitive to the green channel, to improve the perceptual quality, it is possible to increase the samples of green colors by altering the loss function with more emphasis on the green channel. 
They are left as future work. 

\heading{Qualitative comparison.}
\figref{visual_comp} shows the visual results of DMCNN-VD-Pa for several examples.
Compared with the results of DMCNN-VD with the Bayer CFA, the new CFA helps correcting quite a few artifacts. For example, in \figref{visual_comp}(b), the result of DMCNN-VD-Pa is crisper and sharper than the one of DMCNN-VD. 
In \figref{visual_comp}(d), compared with DMCNN-VD, the zipper effect is almost completely removed by the new pattern of DMCNN-VD-Pa. 

\subsection{Demosaicing with spatially varying exposure and color}
\label{sec:SVEC}

\begin{figure}[t]
\begin{center}
\begin{tabular}{ccc}
\includegraphics[width=0.32\linewidth]{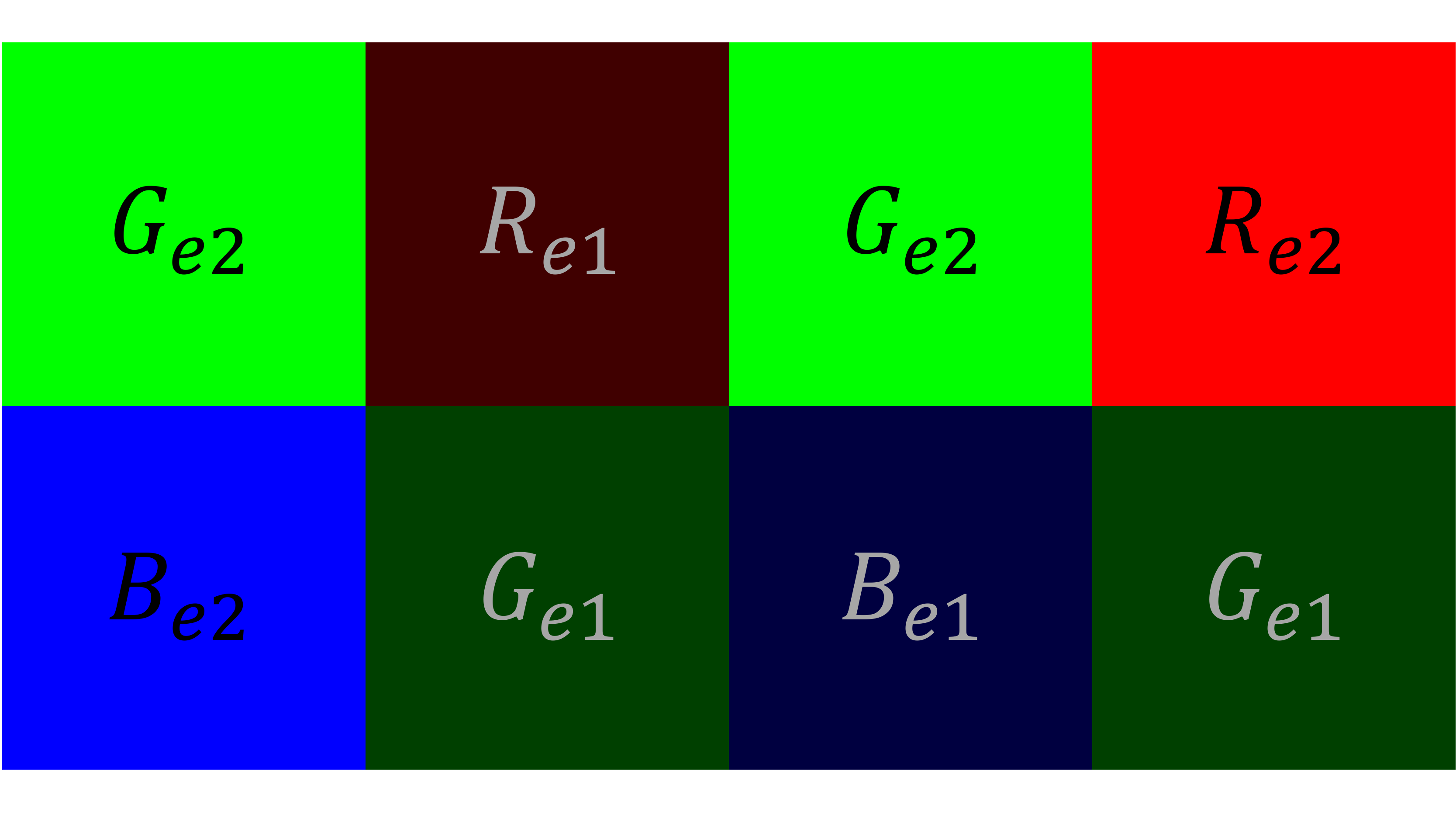} &
\includegraphics[width=0.32\linewidth]{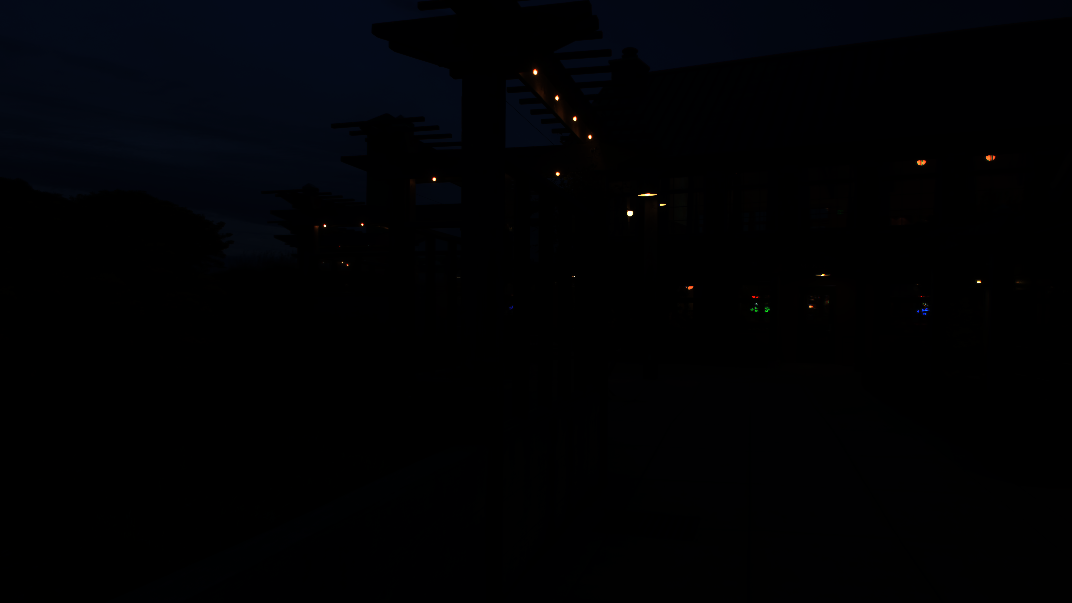} &
\includegraphics[width=0.32\linewidth]{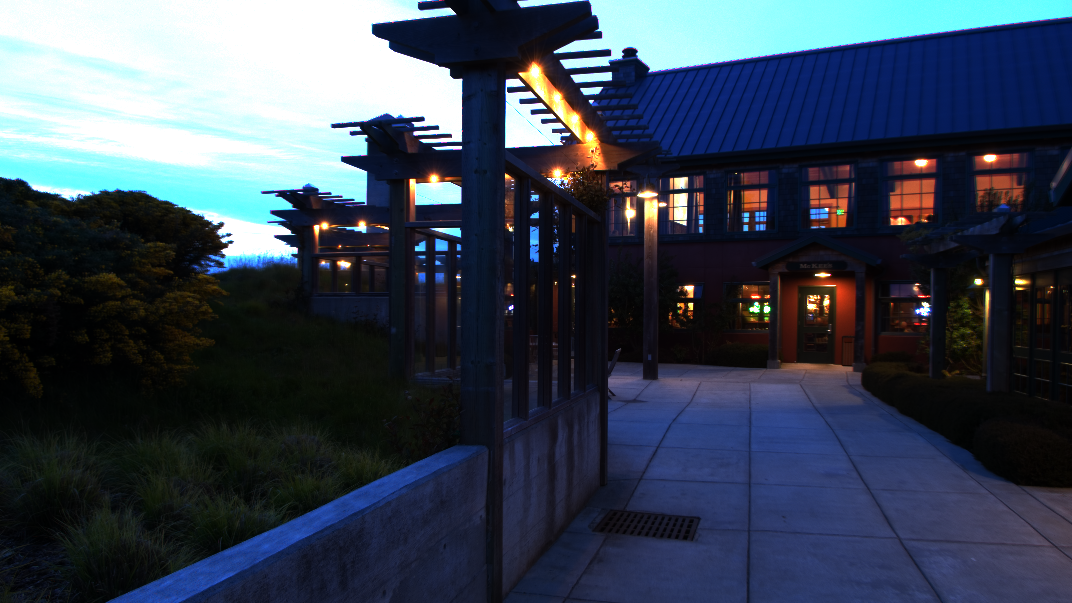} \\
(a) & (b) & (c)
\end{tabular}
\caption{The SVEC configuration. (a) The SVEC pattern used in the HDR experiment. (b) The image captured with the low exposure. (c) The image captured with the high exposure.}
  \label{fig:hdr}
\end{center}
\end{figure}

In addition to color demosaicing, the CNN model can also be applied to more general demosaicing problems. Here, we address the problem of demosaicing with spatially varying exposure and color (SVEC) sampling.
More specifically, the CFA takes samples with different combinations of both colors and exposures.
\figref{hdr}(a) gives a CFA design with three color channels, R, G and B, and two exposures, the low exposure $e_{1}$ and the high exposure $e_{2}$ (the high exposure is 64 times higher than the low one in our setting).
It extends the Bayer CFA with spatially varying exposures. 
\figref{hdr}(b) and \figref{hdr}(c) show the images of the same scene captured with these two exposures.
By taking pictures with the SVEC CFA in \figref{hdr}(a), it is possible to reconstruct a high dynamic range (HDR) image using only a single shot.
However, the SVEC demosaicing problem is more challenging than color demosaicing since there is more information loss in SVEC demosaicing (5/6 of information is lost) than color demosaicing (2/3).
Thanks to the flexible, end-to-end CNN model, we can address the more challenging SVEC demosaicing problem with the same models and proper training data.
In this case, we have six channels in the input and the output is an HDR image with three color channels. 
Note that, rather than reconstructing six channels corresponding \revision{to} RGB colors with two exposures, we directly recover real-valued RGB colors as the output.

\setlength{\tabcolsep}{2pt}
\setlength{\tabcolsep}{6pt}

\heading{Training data.}
For the problem setting of SVEC demosaicing, we need HDR images for simulating the captured images with different exposures.
Unfortunately, HDR images often have quite different ranges and it could be problematic for training CNN models.
To deal with this problem, we normalize the radiance images as
\begin{align}
  \tilde{I}_{i}=\lfloor \min(\frac{r_{max}-r_{min}}{I^{max}-I^{min}}*(I_{i}-I^{min}),r_{max}) \rfloor,
\end{align}
where $\tilde{I}_{i}$ is the normalized radiance for the pixel $i$;
$r_{max}$ and $r_{min}$ are the maximum and minimum radiance values of the output range ($2^{12}$ and $2^{-6}$ respectively in the current setting); $I^{max}$ and $I^{min}$ denote the maximum and minimum values of the original radiance image $I$; the $\min$ and floor function  simulates clamping and quantization of the camera pipeline (in our setting, the simulated sensor has 12 bits per pixel).
After normalization, the SVEC pattern is applied to simulate the input.
We collected 180 HDR images online and divided them into three subsets (100 for training, 30 for validation and 50 for testing) for the following experiments.

\heading{Quantitative comparison.}
For SVEC demosaicing, we compare our \revision{models} to Assorted Pixel (AP) proposed by Nayar and Narasimhan~\cite{nayar2002assorted} using MSE (mean square error) and CPSNR as metrics.
\tabref{HDR_comp} reports the results.
The DMCNN model significantly outperforms AP in both metrics.
\revision{With its deeper architecture, DMCNN-VD further improves the MSE error and the CPSNR value.}
It shows that the CNN models are more powerful than the simple regression model used by AP~\cite{nayar2002assorted}.
In addition, AP cannot capture the spatial relationships as well as the CNN models. 

\begin{table}[t]
\centering
\begin{tabular}{|c|c|c|}
\hline
Algorithm                  & MSE   & CPSNR \\ \hline \hline
AP~\cite{nayar2002assorted} & 54.76 & 42.55 \\ \hline
DMCNN & 28.23 & 47.25 \\ \hline
\revision{DMCNN-VD} & 14.89 & 53.13 \\ \hline
\end{tabular}
\caption{Quantitative evaluation for SVEC demosaicing, in terms of the average MSE and CPSNR values for 50 testing HDR images.}
\label{tab:HDR_comp}
\end{table}

\heading{Qualitative comparison.}
\figref{hdr_visual_comp} shows the SVEC demosaicing results for two testing images.
For each example, we show the ground truth radiance maps and the radiance maps recovered by \revision{AP, DMCNN and DMCNN-VD}, all visualized with the heat map.
The difference maps show that the results of the DMCNN model are closer to the ground truth as it has more blue colors in the difference maps.
\revision{With its deeper structure, DMCNN-VD further reduces the errors.}
The close-ups shows that the DMCNN model generates less artifacts around edges \revision{than AP while DMCNN-VD outperforms DMCNN with even sharper edges}. 

{
\begin{figure*}
\centering
\renewcommand{\arraystretch}{2}

\resizebox{2\columnwidth}{!}{%
\begin{tabular}{C{5cm}C{5cm}C{5cm}C{5cm}}
\toprule[1pt]
GT tone-mapped & diff(AP~\cite{nayar2002assorted}, GT) & diff(DMCNN, GT) & \revision{diff(DMCNN-VD, GT)}\\ \hline
GT radiance & AP~\cite{nayar2002assorted} radiance & DMCNN radiance & \revision{DMCNN-VD radiance} \\ \hline
close-up of GT & close-up of AP~\cite{nayar2002assorted} & close-up of DMCNN & \revision{close-up of DMCNN-VD} \\ \bottomrule[1pt]
\end{tabular}
}
\renewcommand{\arraystretch}{0.5}
\begin{tabular}{ccccc}
&&&&
\end{tabular}

\resizebox{2\columnwidth}{!}{%
\begin{tabular}{C{4cm}C{4cm}C{4cm}C{4cm}}
\includegraphics[width=\linewidth]{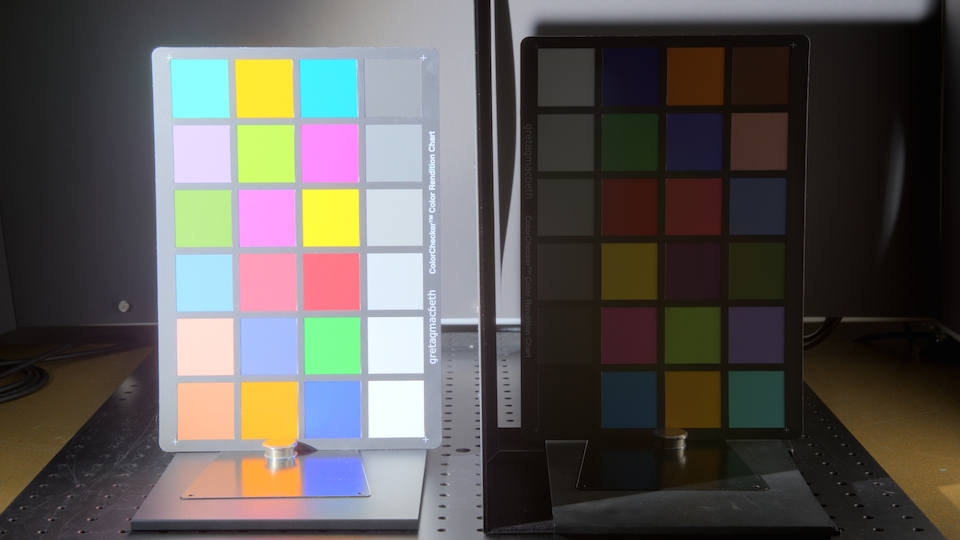}&\includegraphics[width=\linewidth]{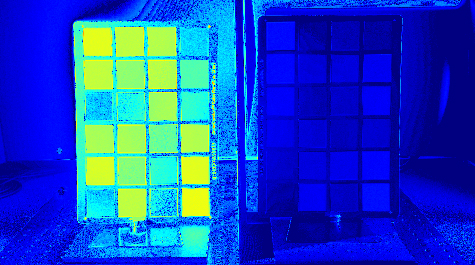}&\includegraphics[width=\linewidth]{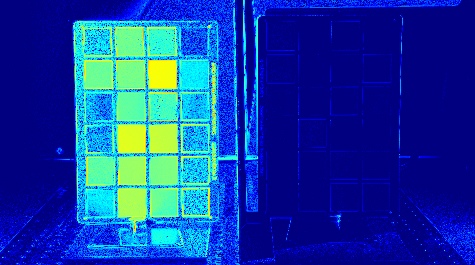}&\includegraphics[width=\linewidth]{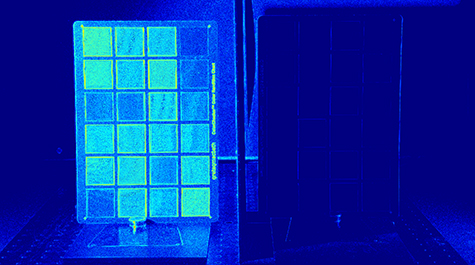}\\
\includegraphics[width=\linewidth]{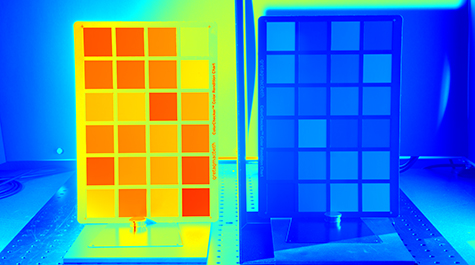}&\includegraphics[width=\linewidth]{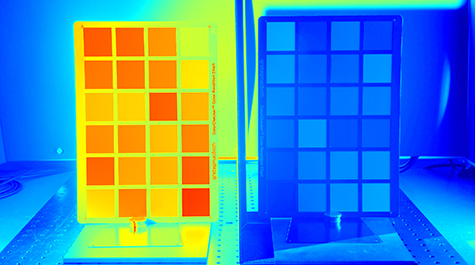}&\includegraphics[width=\linewidth]{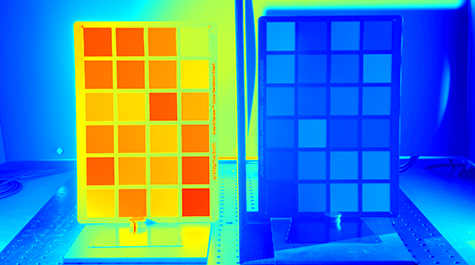}&\includegraphics[width=\linewidth]{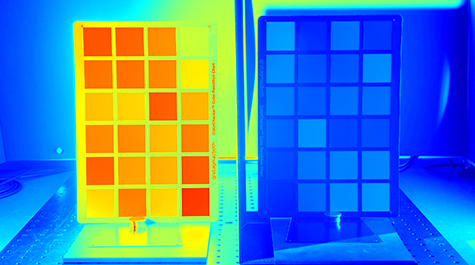}\\
\includegraphics[width=\linewidth]{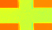}&\includegraphics[width=\linewidth]{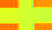}&\includegraphics[width=\linewidth]{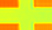}&\includegraphics[width=\linewidth]{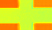}\\
\multicolumn{4}{c}{(a)} \\
\includegraphics[width=\linewidth]{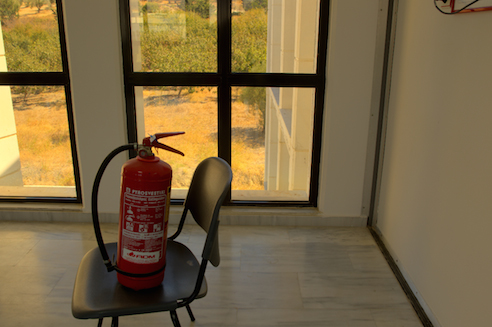}&\includegraphics[width=\linewidth]{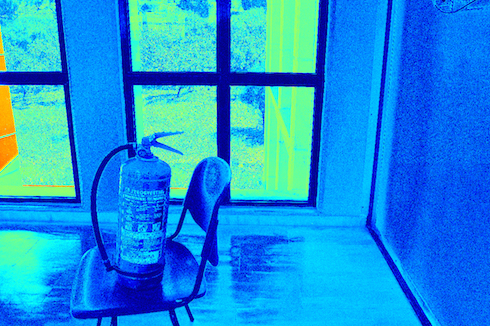}&\includegraphics[width=\linewidth]{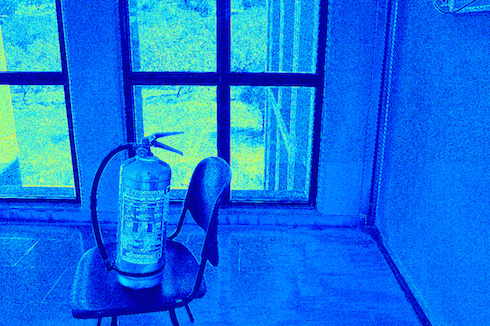}&\includegraphics[width=\linewidth]{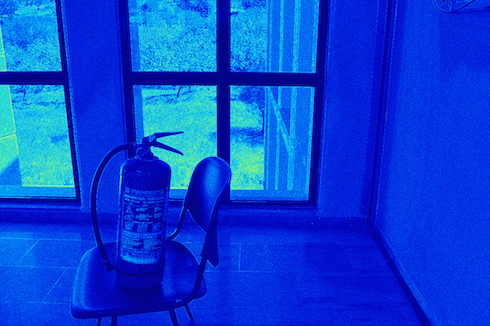}\\
\includegraphics[width=\linewidth]{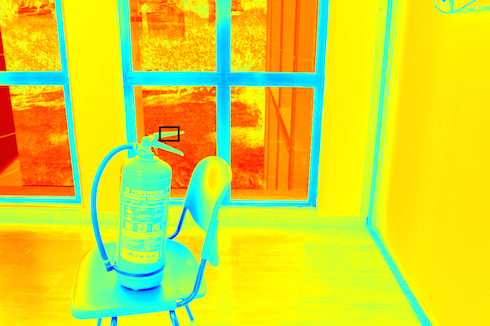}&\includegraphics[width=\linewidth]{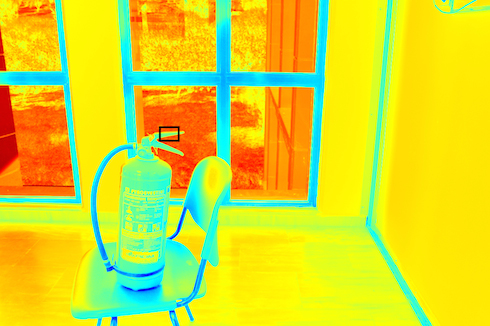}&\includegraphics[width=\linewidth]{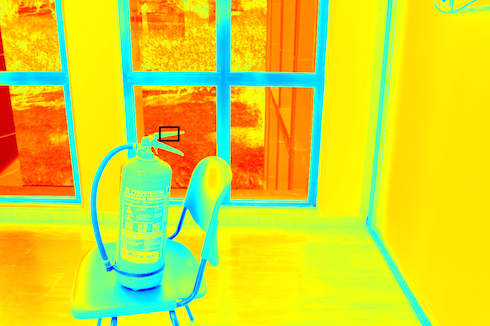}&\includegraphics[width=\linewidth]{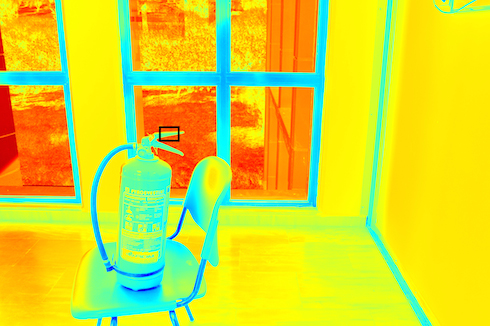}\\
\includegraphics[width=\linewidth]{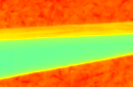}&\includegraphics[width=\linewidth]{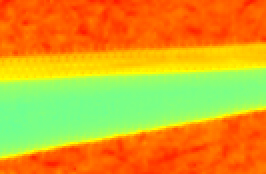}&\includegraphics[width=\linewidth]{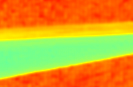}&\includegraphics[width=\linewidth]{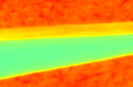}\\ \multicolumn{4}{c}{(b)} \\
\end{tabular}
}
\caption{Visual comparisons on SVEC demosaicing. GT means ground truth. \revision{DMCNN-VD} and DMCNN have less errors as their difference maps contain more blue colors. In general, \revision{both} perform better than AP with less artifacts around edges.}
\label{fig:hdr_visual_comp}
\end{figure*}
}

\section{Conclusions}
\label{sec:conclusion}

In this paper, we present a thorough study on applying the convolutional neural network to various demosaicing problems. Two CNN models, DMCNN and DMCNN-VD, are presented. Experimental results on popular benchmarks show that the learned CNN model outperforms the state-of-the-art demosaicing methods with the Bayer CFA by a margin, in either the sRGB space or the linear space. Experiments also show that the CNN model can perform demosaicing and denoising jointly.  
We also demonstrate that the CNN model is flexible and can be used for demosaicing with any CFA. 
\revision{For example, the current demosaicing methods with the Hirakawa CFA fall far behind the ones with the Bayer CFA.
However, our learned CNN model with the Hirakawa CFA outperforms the-state-of-the-art methods with the Bayer CFA. It shows that the Hirakawa CFA could be a better pattern if a proper demosaicing method is employed.}
It shows the flexibility and effectiveness of the CNN model.
We have also proposed a pattern layer and embedded it into the demosaicing network for joint optimization of the CFA pattern and the demosaicing algorithm. Finally, we have addressed a more general demosaicing problem with spatially varying exposure and color sampling. With the CNN model, it is possible to obtain a high dynamic range image with a single shot. 
All experiments show that the CNN model is a versatile and effective tool for demosaicing problems. 

\ignore{
In this paper, we first present a CNN-based method for the classical demosaicing problem.
The proposed models (DMCNN, DMCNN-DR) show competitive performances compared to state-of-the-art algorithms.
Our DMCNN-DR outperforms most methods on the popular benchmark (Kodak, McM).
We also showed the proposed method can also tolerate linear, noisy data (MMD).
By adopting transfer learning, we showed the fine-tuned model can perform much more better.
Second, we presented a new layer, pattern layer, to simulate the behavior of CFA.
By embedding pattern layer into previously introduced demosaicing CNN, we can optimize pattern design and demosaicing jointly.
Our experiments showed the significant improvements on both PSNRs and visual quality.
Final, we extended the problem into more generalized case, SVEC demosaicing.
The results are also promising and the details in HDR images can be reconstructed well.
We speculate the performance of the proposed method could be further enhanced by combing self-similarity properties into our CNN architecture or as some post-processing.
Overall, the proposed model has achieved the state-of-the-art performance and can be more practical due to its straightforward, non-iterative properties. 
}



\ifCLASSOPTIONcaptionsoff
  \newpage
\fi

\bibliographystyle{IEEEtran}
\bibliography{TIP_demosaicking}



\end{document}